%% file: main.tex
\documentclass[letterpaper]{article} % DO NOT CHANGE THIS
\usepackage{aaai23}  % DO NOT CHANGE THIS
\usepackage{times}  % DO NOT CHANGE THIS
\usepackage{helvet}  % DO NOT CHANGE THIS
\usepackage{courier}  % DO NOT CHANGE THIS
\usepackage[hyphens]{url}  % DO NOT CHANGE THIS
\usepackage{graphicx} % DO NOT CHANGE THIS
\def\UrlFont{\rm}  % DO NOT CHANGE THIS
\usepackage{natbib}  % DO NOT CHANGE THIS AND DO NOT ADD ANY OPTIONS TO IT
\usepackage{caption} % DO NOT CHANGE THIS AND DO NOT ADD ANY OPTIONS TO IT
\usepackage{algorithm}
\usepackage{algorithmic}
\usepackage{amsmath}
\usepackage{amsfonts}
\usepackage[export]{adjustbox}
\usepackage{multirow}
\usepackage{tikz}
\usetikzlibrary{shapes,chains,scopes,positioning,arrows}
\tikzset{block/.style={draw, rectangle, rounded corners, thick}}
\usepackage{booktabs}
\usepackage{newfloat}
\usepackage{listings}
\DeclareCaptionStyle{ruled}{labelfont=normalfont,labelsep=colon,strut=off} % DO NOT CHANGE THIS
\floatstyle{ruled}
\newfloat{listing}{tb}{lst}{}
\floatname{listing}{Listing}
\title{Show, Interpret and Tell: Entity-aware Contextualised Image Captioning in Wikipedia}
\author {
    Khanh Nguyen,
    Ali Furkan Biten, 
    Andres Mafla,
    Lluis Gomez,
    Dimosthenis Karatzas
}
\affiliations {
    Computer Vision Center, Universitat Autònoma de Barcelona, Barcelona, Spain\\
    \{knguyen, abiten, andres.mafla, lgomez, dimos\}@cvc.uab.es
}
\usepackage{bibentry}
\begin{document}

\maketitle

\begin{abstract}
Humans exploit prior knowledge to describe images, and are able to adapt their explanation to specific contextual information, even to the extent of inventing plausible explanations when contextual information and images do not match. In this work, we propose the novel task of captioning Wikipedia images by integrating contextual knowledge. Specifically, we produce models that jointly reason over Wikipedia articles, Wikimedia images and their associated descriptions to produce \textit{contextualized} captions. Particularly, a similar Wikimedia image can be used to illustrate different articles, and the produced caption needs to be adapted to a specific context, therefore allowing us to explore the limits of a model to adjust captions to different contextual information. A particular challenging task in this domain is dealing with out-of-dictionary words and Named Entities. To address this, we propose a pre-training objective, Masked Named Entity Modeling (MNEM), and show that this pretext task yields an improvement compared to baseline models. Furthermore, we verify that a model pre-trained with the MNEM objective in Wikipedia generalizes well to a News Captioning dataset. 
Additionally, we define two different test splits according to the difficulty of the captioning task. We offer insights on the role and the importance of each modality and highlight the limitations of our model. The code, models and data splits are publicly available at ~\url{Upon acceptance}.
\end{abstract}

\section{Introduction}
% Human-level 
Scene understanding involves composing a story that explains the perceptual observations. This ability, to draw from previous experience to explain what is happening, is termed image interpretation~\cite{lake2017building}.
% Human-level scene understanding involves composing a story that explains the perceptual observations with image interpretation at play~\cite{lake2017building}. 
Image interpretation is one of the hallmarks of linguistic intelligence~\cite{gardner2011frames} where contextual information is drawn upon to compose an explanation of the depicted events~\cite{terman1916measurement}. 
We identify Wikipedia as an excellent probing ground to further the advance in image interpretation and  integrate contextual information into captioning models.

\begin{figure}[t!]
    \centering
    %  View from Arni to Chli Windgällen (2986m, in front) and the Groß Windgällen (3188m, behind) in the Canton of Uri/Switzerland
    \includegraphics[width=0.48\textwidth]{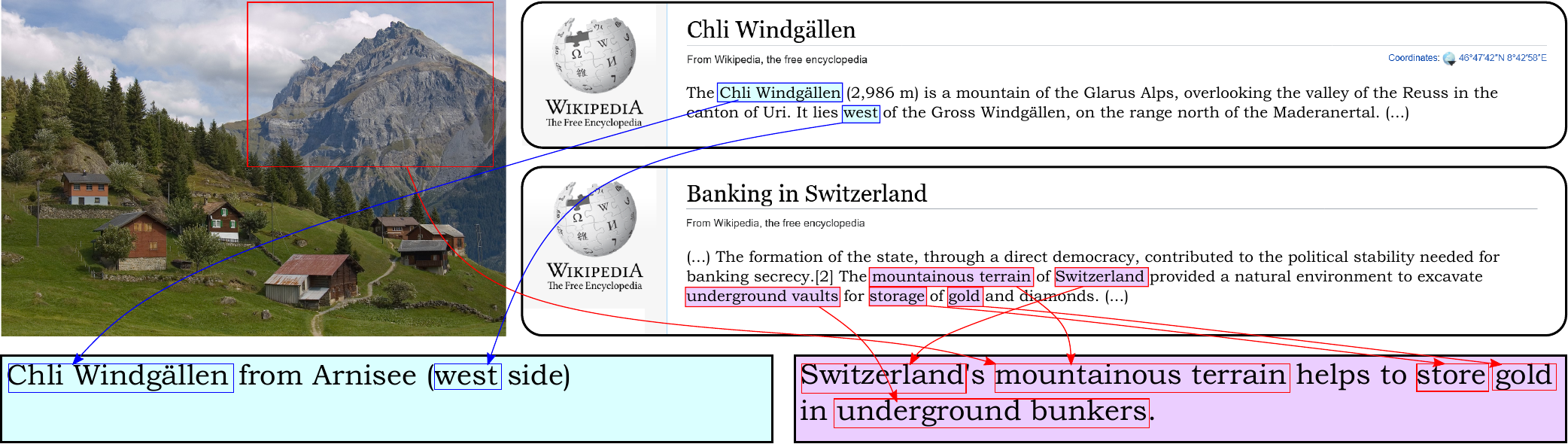}
    \caption{\textbf{The role of contextual information in Wikipedia Captioning.} The same image can have completely different captions depending on contextual information given in the form of article and section. On the top side, the corresponding page is a description of a location, while the bottom one talks about Banking in Switzerland.}
    \label{fig:teaser}
    \vspace{-0.45cm}
\end{figure}

Wikipedia is an invaluable source of world knowledge information. Each page consists of different information sources comprised of title, sections, images and their captions. Each author of Wikipedia writes an article and illustrates it with images that are selected from Wikimedia. Each image has a corresponding description that includes very detailed information, including Named Entities (NEs). In order to write a good caption for the image while remaining relevant to the article, the writers need to take into account the section they wrote, the image they selected and its corresponding description to produce a contextualized caption, a task referred to as \textit{Wikipedia captioning}. In this work, 
% and to the best of our knowledge, 
we propose Wikipedia Captioning as a real use case scenario to alleviate the work load of Wikipedia
writers while allowing them to better illustrate their articles. Wikipedia captioning offers the unique opportunity to potentially aid thousands of writers directly and millions of readers indirectly on a daily basis. 
% Nonetheless, this task is easily transferable to illustrate and caption images relevant to a specific context automatically.

To grasp the complexity of Wikipedia captioning, an example of a contextualized caption can be observed in Figure~\ref{fig:teaser}. The importance of the context is paramount given the possibility of having several matching and distinct captions for a single image. Naturally, this introduces numerous challenges. On one hand, the context (description and section) needs to be encoded and information selectively drawn from it while being guided by the visual content. On the other hand, explicit contextual information, typically found in the form of NEs such as proper names, prices, locations, dates, etc., needs to be properly injected in the generated captions in natural language. This explicit contextual-dependant information is typically out-of-dictionary or at best it is underrepresented in the
statistics of the dictionary used. To better prepare a model that deals with Named Entities, we propose a pre-training strategy in which we extend Masked Language Modeling (MLM)~\cite{devlin2018bert} to Masked Named Entity Modeling (MNEM). We show that MNEM is more effective at selecting the correct NE given an article while reducing the language prior produced by NEs.    

% Relevance to news captioning
Wikipedia Captioning is closely related to News Captioning~\cite{biten2019good,feng2013automatic,ramisa2018breakingnews}. In the case of News Captioning, a single source of context is available in which the news article is roughly equivalent to a Wikipedia article, and no separate (article independent) description for the image is available. Yet in Wikipedia Captioning, images are associated to specific sections of the article, plus specific article-independent descriptions for each image are also given. The two tasks would be equivalent if we ignore the descriptions and we equate a Wikipedia section with the whole news article.
% All aforementioned challenges are realized in news captioning~\cite{biten2019good,feng2013automatic,ramisa2018breakingnews}. Given the similarities of the task between Wikipedia and News captioning\footnote{Wikipedia captioning conceptually is the same as News captioning when we assume we do not have any description of an image and treat the section as the article for the image.}, 
Given the similarities between the two tasks, we examine whether the performance can be improved by transferring from Wikipedia to News captioning. We discover that the knowledge acquired by the model on Wikipedia captioning is transferable to News captioning, achieving better results over the baseline. Moreover, we demonstrate that MNEM is effective as well on News captioning, outperforming a base model in terms of automatic image captioning evaluation metrics.

Furthermore, we uncover several key insights on the Wikipedia captioning task and on the capabilities of the proposed methodology. Firstly, since such models are prone to take shortcuts~\cite{geirhos2020shortcut}, we discover a bypass strategy that our model is taking, a copying process of NEs from descriptions to captions. 
% , and where we explore in depth the effect this has. 
Additionally, humans have the ability to come up with a caption for any image given a specific context (article, section, description, etc.). Contrary to open-world story generation~\cite{martin2018event} which is to come up with stories for any topic without prior knowledge, context-based story-telling is to generate captions with context acting as a prior knowledge for any given image. Trying to emulate this specific skill, we examine if our model can generate a contextualized caption given mismatched but potentially relevant images. 
% coming from a retrieval system. 
Despite our model's state-of-the-art performance according to captioning metrics and injecting the most relevant NEs, we reveal that current models are still far off from endowing this skill. To summarize, the key contributions of our work are: 
\begin{itemize}
    \item We introduce the Wikipedia Captioning as a real use case scenario to benefit writers and readers on a daily basis. 
    \item We propose an architecture-agnostic pretext task, namely Masked Named Entity Masking, that can be applied to any contextualized captioning model. We empirically show that when it is used, a gain in performance is achieved on Wikipedia and News captioning domains.
    \item We exhibit that pre-training models on Wikipedia captioning boosts performance on News captioning, transferring the knowledge from one domain to another.  
    \item We extensively study the limitations of our models and discuss a setting that shows we are far off on mimicking the human ability of context-based story-telling. 
\end{itemize}

\section{Related Work}
\textbf{Classic Image Captioning.} 
Early image captioning methods~\cite{farhadi2010every,ordonez2011im2text} focus on hand crafted features with templates. The advent of deep learning with large scale datasets~\cite{lin2014microsoft}, in contrast, steers the trend into end-to-end encoder-decoder networks. Centralized by attention mechanisms~\cite{Bahdanau2014,devlin2018bert}, architectures, ranging from LSTM-based~\cite{vinyals2015show,You,anderson2017bottom,lu2017knowing} to Transformers~\cite{pan2020x,luo2021dual,ji2021improving,chen2021captioning} achieve remarkable progress in captioning metrics. Advanced models~\cite{dai2017towards, shetty2017speaking} employ GANs to favor human-like caption outputs. Works applying Reinforcement Learning~\cite{Rennie2016,liu2017improved} successfully optimize models on non-differential metrics. Nevertheless, such models are restricted by design to the mere description of what is shown, thus unable to attempt any interpretation.

\textbf{News Articles Captioning.} Recent advances in News image captioning~\cite{feng2013automatic} come in both architectures and datasets. Typically, proposed models follow the standard encoder-decoder paradigm while large-scale datasets such as Breaking News~\cite{ramisa2018breakingnews}, GoodNews~\cite{biten2019good}, VisualNews~\cite{liu2020visual} etc. have been contributed. To deal with NEs, ~\cite{biten2019good, liu2020visual} propose two-stage template-based methods that train models to capture NE-tag distribution followed by post-generation NE insertion. Closely related to our work, ~\cite{tran2020transform} employs a transformer~\cite{vaswani2017attention} with byte-pair-encoding (BPE)~\cite{sennrich2015neural} to directly decode NEs. Lastly,~\cite{yang2021journalistic} enhances caption quality with template guidance and a pre-trained NE Embedding module based on Wikipedia Knowledge graph~\cite{yamada2018wikipedia2vec} to better encode NEs. Given the importance of NEs in this task, we introduce the Named Entity Masking pretext task for accurate NE prediction and coverage.
\textbf{Self-supervised Pre-training Methods}.
Self-supervised approaches in NLP~\cite{devlin2018bert,radford2019language,roberts2020exploring} have greatly inspired many works in vision-and-language (VL)~\cite{li2020oscar,kim2021vilt,zhang2021vinvl} that follow the pre-training and fine-tuning paradigm, elevating the performance in a wide variety of benchmarks.
While existing pre-training objectives in VL either focus on representation learning or multi-modal alignments, our proposed method MNEM aims to learn context-entity alignment via the MLM protocol. 
Compared to other entity-centric methods~\cite{sun2020ernie, lin2021entitybert}, we are the first to explore MNEM in a multi-modal scheme, with careful adaptation to suit the image-text distribution found in Wikipedia.
\section{Method}
In this section, we first define the Wikipedia captioning task formally and then present our approach which utilizes well-designed transformer-based models to solve the task.
%\textit{modified} Transform and Tell (M-TaT) based model. 
We further discuss how we extend these models by employing a novel pre-training objective which we refer to as Masked Named Entity Modeling (MNEM).
\begin{figure*}
\centering
\begin{minipage}{0.48\textwidth}
    \centering
    \includegraphics[width=0.95\linewidth, height=0.25\textheight]{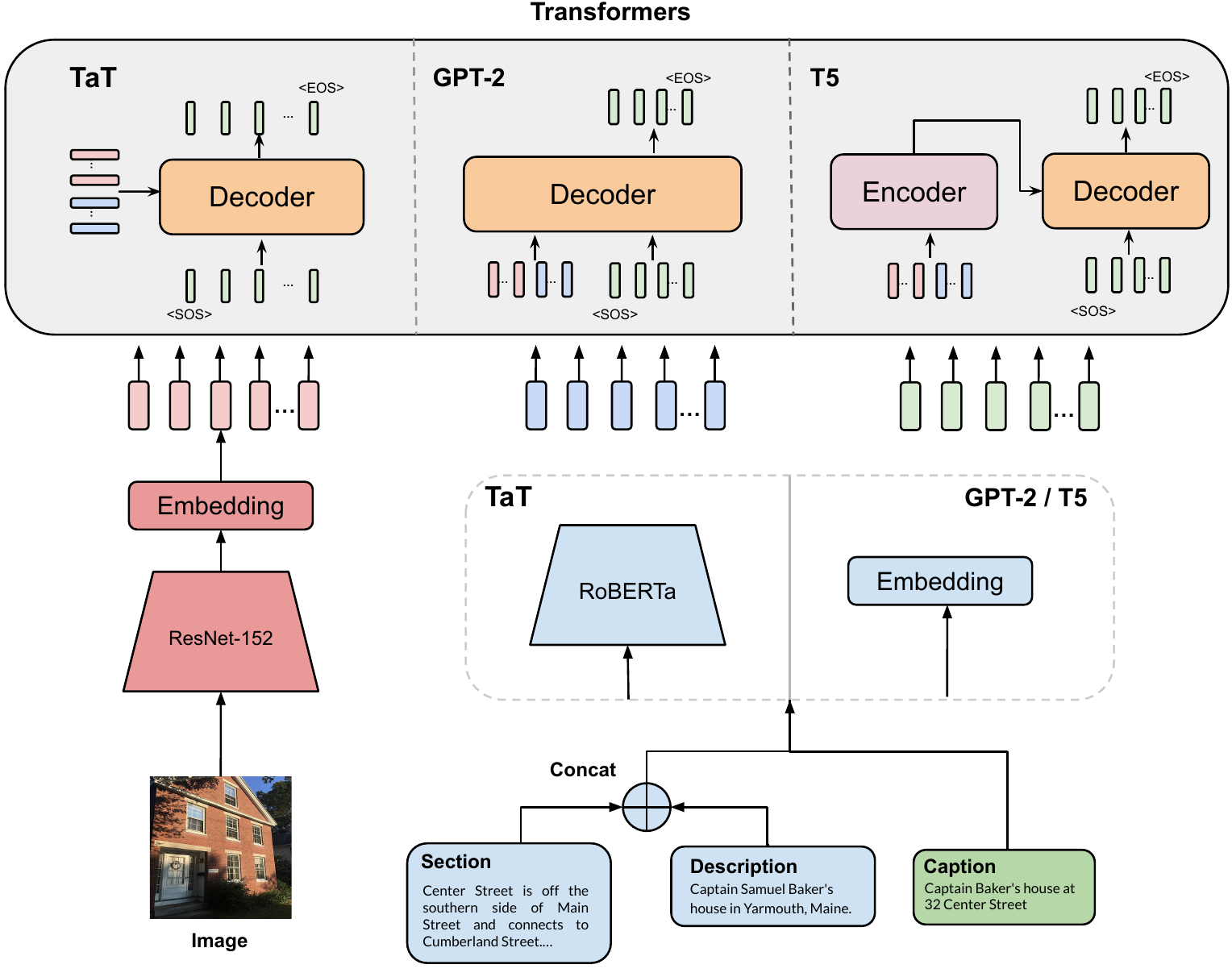}
    \end{minipage}%
\unskip\ \vrule\
\begin{minipage}{0.48\textwidth}
    \centering
    \includegraphics[width=\linewidth, height=0.25\textheight]{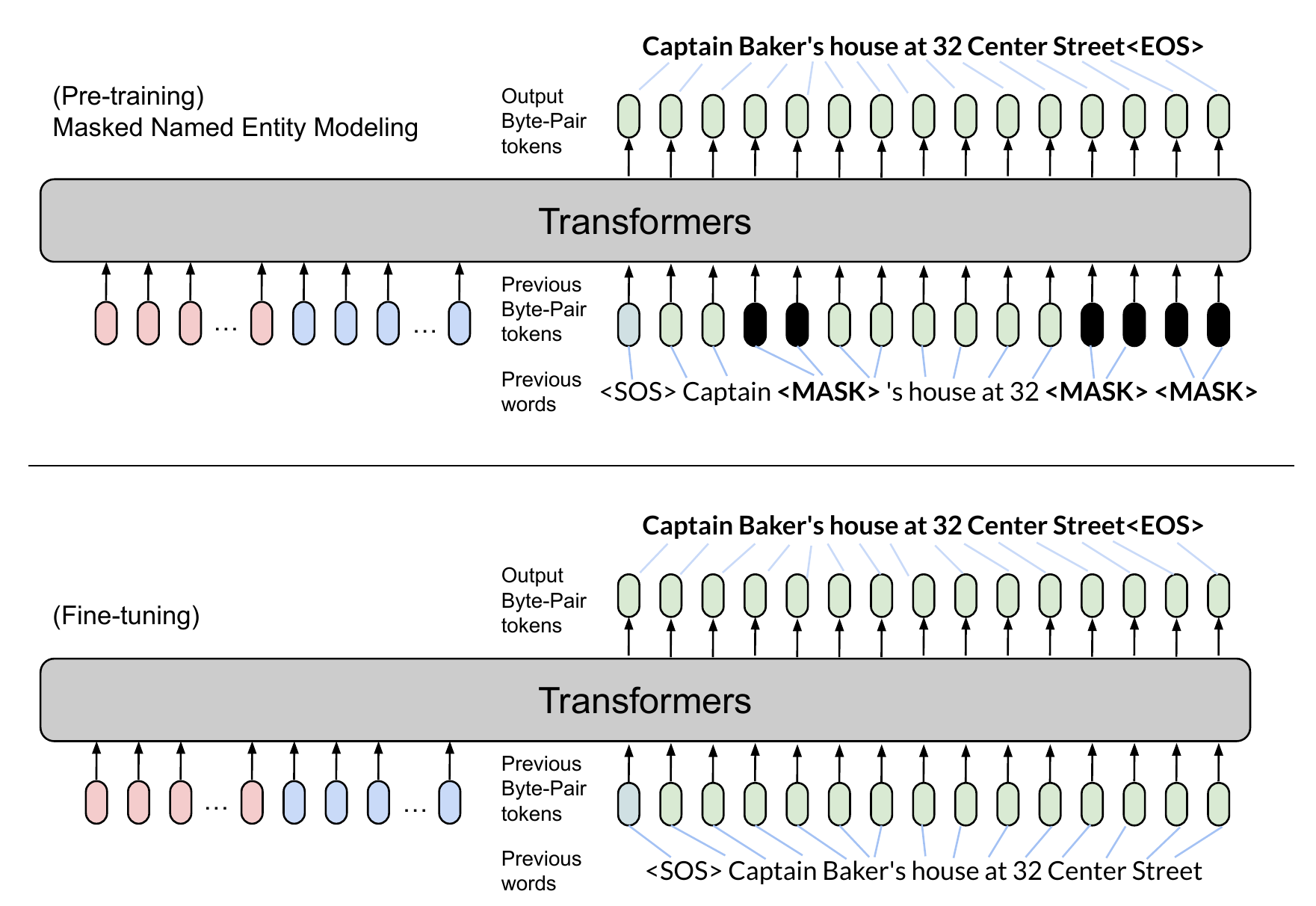}
        % \caption{Jaccard similarity of caption and section}
        % \label{fig:prob1_6_1}
\end{minipage}
\caption{\textbf{Overview of the Transformer-based models and our MNEM pre-training task.} Left: Transformers take multi-modal input features (image, section, description) processed by single-modal encoders; Right: During pre-training (right-top) we replace the NEs in the ground-truth caption by the \textit{MASK} token and train our models to reconstruct the caption.
% By doing so, the model is able to learn better representations which are useful for the  Wikipedia caption prediction task.
}
\label{fig:tat}
% \vspace{-.5cm}
\end{figure*}

\subsection{Problem Definition}
% We start by formally defining the Wikipedia captioning problem. 
We refer to $I=\{i_1, i_2, ...\}$, $C=\{c_1, c_2, ...\}$ as the image set and its corresponding captions set where each caption contains words defined as $c_{k_j}$ or $c_k = \{c_{k_0}, c_{k_1}, ...\}$. Moreover, $D=\{d_1, d_2, ...\}$ and $S=\{s_1, s_2, ...\}$ refer to description and section sets, respectively. Given a Wikipedia article, a description $d_k$ is tied to each specific image $i_k$ for which it provides Named Entities (when and where the photo was taken, who is depicted in the photo, author, etc.). Each Wikipedia article is divided in sections $s_k$ to enhance readability. To emulate the process that the Wikipedia writers go through during Wikipedia captioning, we define the dataset with quadruples where $(i_k, c_k, s_k, d_k)$ refers to the $k^{th}$ datum. We define the task of generating a caption that is contextualized as $\mathcal{P}(c_k|i_k, s_k, d_k)$.
% \begin{equation}
% \mathcal{P}(c_k|i_k, s_k, d_k)
% \end{equation}
\subsection{Transformer-based Models}
The overview of our proposed method is depicted in Figure~\ref{fig:tat}, which follows the commonly used encoder-decoder paradigm. In particular, we investigate the capacity of Transformer-based models, either trained from scratch or initialized with pre-trained weights acquired from self-supervised training. 
For the latter, we explore Pre-trained language models (PLMs) due to its compelling performances on numerous open-ended text generation tasks.
However, as only pre-trained on textual modality, properly adapting PLMs to solve VL tasks remains a challenge. Therefore, we propose a simple strategy to enable PLMs to reason on multi-modal context when generating captions.
% demonstrating the benefits of large scale self-supervised training that equip these models not only with an impressive ability to generate language but also a rich semantic and linguistic knowledge.
% Therefore, we extend its applicability in the multi-modal setting by including visual information as model's input and 

\noindent
\textbf{Transform and Tell (TaT)~\cite{tran2020transform}. } Proposed to address the News Image Captioning task in an end-to-end manner, TaT consists of a set of frozen encoding modules followed by a trainable Transformer-decoder. The decoder thus generates captions conditioned on a concatenation of multi-modal input features extracted from the image and the article. Perceiving the similarities between the two tasks, we apply TaT to Wikipedia captioning and modify it by stripping away some of the building blocks of the main model. Particularly, we remove the MTCNN~\cite{zhang2016joint} for face detection, FaceNet~\cite{schroff2015facenet} for face embedding, YOLOv3~\cite{redmon2018yolov3} for object representation and the location embedding. Removing these blocks then speeds up the training process by decreasing complexity, while the remaining modules could be easily adapted as required. 
% We also remove this pre-trained components to better assess the proposed model captioning capabilities from a common starting pipeline.   
% Each training sample represents a triplet of image, text, caption (I, T, C).

The reduced encoder now comprises two pre-trained modules for image and text to produce a set of representations. More specifically, given an image $i_k$, we extract image features using a ResNet-152 model pre-trained on ImageNet to obtain ($f_{i_k}$). We use RoBERTa~\cite{liu2019roberta} to encode text sequences into semantic representations capturing a bi-directional context, yielding two sets of text features for sections and descriptions ($f_{s_k}, f_{d_k}$) respectively. The auto-regressive decoder
% , denoted as $g_{\theta}$,
consists of the same BPE embedding matrix introduced in RoBERTa, followed by a stack of $N$ identical transformer blocks. Making use of a shared BPE embedding layer implicitly grants the model the ability of generating rare words. For each transformer block, we use a Dynamic Convolution (DC)~\cite{wu2019pay} layer for past-token conditioning and a Multi-Head Attention (MHA)~\cite{vaswani2017attention} layer for multi-modal conditioning. This architecture design serves as an efficient alternative to the traditional self-attention mechanism, relaxing memory requirements which are quadratic with the input sequence length. At time step $t$, previously-generated token representations $Z_{\le t} = \{z_0, z_1, ..., z_t\}$ are efficiently enriched by attending to its leftward context thanks to the DC. 
% Next, the concatenation over feature sets ($f_{i_k}$, $f_{s_k}$, $f_{d_k}$) is injected to MHA layer to perform cross-attention between the updated partial generation and input contexts. The output of this module $d^{\prime}_{k}$ is used as input to a feed-forward layer with Layer Normalization to before entering the subsequent block. 
After going through the decoder stack, the final representation $z^{\prime}_{t} \in Z^{\prime}_{\le t}$ is fed to the vocabulary layer to predict the next token at $t+1$.

\noindent
\textbf{GPT-2~\cite{radford2019language}} is a transformer decoder variant which performs text generation in auto-regressive manner. 
% Pre-training GPT-2 involves optimizing its weights via a language modelling objective on a great diversity of the dataset. 
% From this perspective, we view Wikipedia captioning as text generation task given the contexts as a prompt. 
We want to leverage the linguistic knowledge stored in the model's pre-trained weights to emulate its success in generating captions, while preserving the Wikipedia style. We therefore decide to fine-tune the entire network, including the learned embedding matrix. We start by transforming the description and section sequences into continuous embeddings ($f_{s_k}$, $f_{d_k}$) using the internal embedding layer of GPT-2. Concurrently, we represent the image as a set of visual features ($f_{i_k}$) extracted from ResNet-152, followed by a MLP layer with Layer Normalization to map images into the word embedding space.
%We call this sequence a visual prefix since it plays the same functional role in the transformer architecture as (part of) an embedding sequence of prefix tokens. 
We form this set of representations ($f_{i_k}$, $f_{s_k}$, $f_{d_k}$) as prefix and pass it to the decoder (referred as $\text{GPT-2}_{\textit{BASE}}\texttt{++}$) whose output $z^{\prime}_{t} \in Z^{\prime}_{\le t}$ is used as a logit vector over the vocabulary distribution to predict the caption token at $t+1$.

\noindent
\textbf{T5~\cite{roberts2020exploring}.}
Built on top of the vanilla Transformer encoder-decoder, T5 is another language model pre-trained on C4. 
% The overall network contains two stacks of transformer blocks, where each block consisting of a self-attention layer followed by a fully-connected layer with residual connections for the encoder and an additional cross-attention layer in the decoder.
% (We suggest readers to check ~\cite{roberts2020exploring} for more details of the model)
Particularly, T5 presents a new self-supervised training objective called \textit{Replace Corrupted Spans} for efficient computation conforming to its encoder-decoder design. To apply the off-the-shelf version of T5 to Wikipedia captioning, we extend its text-only encoder to multi-modal encoder, referred as $\text{T5}_{\textit{BASE}}\texttt{++}$. We incorporate image features $f_{i_k}$ as additional input to the description and section embeddings ($f_{s_k}$, $f_{d_k}$), with each set of embeddings computed the same way as in the previous section. The decoder then attends to previous caption tokens $Z_{\le t}$ via self-attention and the contextualized joint representations from the encoder via cross-attention to predict the next token.

\subsection{Masked Named Entity Modeling}
Referring to the right NEs is among the most important but challenging tasks of Wikipedia captioning to assure the veracity of information included in a caption. MLM, as proposed in BERT~\cite{devlin2018bert}, has shown remarkable success, building richer representations by exploiting the distributional hypothesis present in language to reconstruct masked out words. To extend the applicability of MLM to our captioning task and mitigate the rare NEs insertion issue, we apply the masking strategy explicitly on NEs
% We hypothesize that by masking NEs, we 
to learn more expressive representations for them. More formally, let $\mathcal{W} = \{w_1, w_2, ..., w_n\}$ be the set of all byte-pair tokens. Now, let $\mathcal{M}_l = \{j, j+1, ..., j+k\}$ be the $l^{th}$ mask span where $j$ is the starting index to mask such that $\max(\mathcal{M}_l) < \min(\mathcal{M}_{l+1})$. Then, $\{w_j, ..., w_{j+k}\}$ are replaced with the special mask token [MASK] to form the corrupted caption $\Tilde{\mathcal{W}} = \{\Tilde{w}_1, \Tilde{w}_2, ..., \Tilde{w}_n\}$.
We use cross-entropy to reconstruct the original caption $\mathcal{W}$, with \textit{no restriction to masked tokens} compared to BERT, in order to retain the generation ability:
\begin{equation}
\begin{split}
    \Tilde{w_i} &= \ [MASK], \text{where} \ i\in \mathcal{M}_l \\
    \mathcal{L} &= \sum_{i=0}^{N}{-\log{p(w_i|\Tilde{\mathcal{W}}_{0<i}, f_{i_k}, f_{s_k}, f_{d_k})}}
\end{split}
\label{eq:masking}
\vspace{-3cm}
\end{equation}
Inspired from T5, we introduce a small modification to MNEM to specifically pre-train encoder-decoder architectures. We first substitute each entity span $\mathcal{M}_l$ with \textit{a unique mask} token to create $\Tilde{\mathcal{W}}$ then append it to the encoder input sequence. Consequently, instead of reconstructing the caption, we task the decoder to predict all the masked name entities, which are concatenated in original order as the target sequence. This conforms the training procedure performed on T5 while explicitly tells the model to focus on name entities within context.

% \noindent
% \begin{figure}
% \centering
% \includegraphics[height=2.5cm]{masked_NE.png}
% \caption{The input caption during pre-training is modified by replacing major NEs with mask elements. By doing so, the model is able to learn better representations which are useful for Wikipedia caption prediction task.}
% \label{fig:masked_ne}
% \end{figure}
%masking strat
Figure~\ref{fig:tat} depicts our proposed design with masked NEs language modeling during the pre-training stage. We extract NEs contained in captions
% and replace them with the [MASK] token. 
% We follow the implementation details of BERT~\cite{devlin2018bert} where 
and randomly mask the NEs with a probability of 0.8 while adopting a whole word masking strategy. We only mask NEs that fall into the categories of: [\textit{Person}, \textit{ORGanization}, \textit{GeoPolitical Entity}] as they appear to be dominant types in Wikipedia captions. Moreover, changing every NEs might be too disrupting of the semantic structure in captions which tend to be short. 
% The decoder is trained to reproduce the original input caption shifted one step to the right, while the main structure remains unchanged, similarly to the one used by our M-TAT. 
% The auto-regressive operation is effectively governed by the MSA [] in which the decoder can only look at earlier tokens, both masked and non-masked, to generate the next one. 
%advantages
% Unlike MLM, 
% we do not decouple the masked NEs prediction from the original decoding task. Consequently, 
% the objective does not restrict the prediction only on masked tokens but instead we predict all tokens to maintain the generation ability. At each time step, predicted tokens are conditioned on the leftward context with multi-modal inputs. 
% Moreover, 

By masking the NEs within the captions, the model can learn to precisely capture the %right
contextual information relating to the NEs to produce useful representations. 
Also, our proposed method aims to mitigate 
% two issues: First, by masking the NEs from the model, we reduce the influence of the Named Entity bias inherently included in the training data. NEs that do co-occur frequently tend to have higher probabilities to be generated regardless of a given context, thus decreasing the model's ability to select correct NEs.
%Second, as shown in later sections, in cases 
another issue, which is further studied below. 
We found that when there is a high overlap between the captions and context, the model learns to directly copy NEs to generate captions, making it ineffective to deal with the diversity and semantic richness of Wikipedia Captioning. By limiting the input caption with non-NEs words in training phase, we decrease the influence of the language bias, allowing a better utilization of the entire context given as input.

\section{Experiments}
In our experimentation for Wikipedia Captioning, we utilize the WIT~\cite{srinivasan2021wit} dataset. All the details regarding the dataset statistics, pre-processing and implementation as well as an in-depth explanation of our baselines can be found in the Appendix.

\input{tables/wiki_cap}
\subsection{Wikipedia Captioning}
Table~\ref{table:wiki_2_cap} showcases the results for Wikipedia Captioning task in BLEU-4~\cite{Papineni2002}, METEOR~\cite{banerjee2005meteor}, ROUGE-L~\cite{lin2004rouge}, CIDEr~\cite{vedantam2015cider}, and SPICE~\cite{anderson2016spice}, as well as precision and recall of the NE insertion as defined by~\cite{biten2019good}. The high performing baselines especially without image imply strong language priors learned by the models, outperforming transformer language models like TaT or GPT-2. The observed drawbacks of such methods indicate two things: (1) the baselines tend to rely on copying NEs excessively coming from the description of the images only and (2) the models prefer to generate few NEs. These hypotheses are confirmed by the high Precision and low Recall, suggesting the models have found a shortcut to generate few NEs by only copying them from the description. We examine this behaviour more thoroughly in the analysis section. Moreover, the addition of images as an input clearly disrupts the performance. According to our examination, we find out that the attention weights of seq2seq models are significantly lower for the image modality than the context one. In other words, seq2seq tries to disregard the images while putting more emphasis on the context, failing to find the correspondence between two modalities. 

Yet, we see opposite behaviour on the efficacy of images on transformer models, showcasing the superiority of transformers over LSTM encoder-decoder. The addition of images significantly improves the performance on all the captioning metrics. Interestingly, images have a direct impact on the Recall for TaT, improving the results from \textbf{27\%} to \textbf{34\%} while not affecting the Precision score. Improving the Recall score discerns a crucial lesson that aligning the image and text modality is imperative to generate higher number of correct NEs. However, this effect is not present in GPT-2 and T5, which is expected, as these PLMs naturally gravitate towards linguistic features, thus require more well-designed adaptation to balance the visual and textual information in VL tasks. MNEM aims to connect two modalities by means of predicting NEs, leading to better NEs selection given multi-modal alignment. It is clearly observed that in all transformer models, MNEM exceeds the performance significantly on captioning metrics (\textbf{+10} and \textbf{+5} points in CIDEr for GPT-2 and TaT respectively). MNEM also provides cues to select NEs properly, gaining on both Precision and Recall with or without image. Finally, T5 shows a consistent superiority among all models, with MNEM it achieves state-of-the-art in Wikipedia Captioning task.
\input{tables/goodnews}
\subsection{Transfer Learning from Wikipedia to News}
Given the similar challenges that the two tasks face, we examine the possibility of transfer learning from Wikipedia to News captioning, \textit{i.e.} fine-tuning models on GoodNews with weights \textbf{initialized from pre-training on Wikipedia}. To imitate the News captioning setting, we utilize the section content as the only context, disregarding the description. We select the  $\text{T5}_{\textit{BASE}}\texttt{++}$ model for the experiments given that it is the best performing model in Wikipedia Captioning. 
Table~\ref{table:goodnews} shows the results of the transferability from Wikipedia to News with MNEM pre-training. At first, simply fine-tuning $\text{T5}_{\textit{BASE}}\texttt{++}$ on GoodNews can match or surpass previous transformer models (\textbf{+6} points in CIDEr compared to TaT), demonstrating PLMs impressive ability to generate captions. Pre-training on Wikipedia further improves the performance over JoGANIC, which is the state of the art method with Wikipedia pre-training, (\textbf{+0.36} points in BLEU-4 and \textbf{+2.67} in CIDEr) while not including specific template guidance. This illustrates the advantage of proposing this \textbf{Wikipedia Captioning} task, in which the acquired knowledge is useful for the News domain. Moreover, incorporating MNEM pre-training on GoodNews constantly results in additive gains, demonstrating its effectiveness in this task. However, the improvement is not transferable as we additionally perform MNEM on Wikipedia (rows 9–12). This extra pre-training stage guides the model to focus on Wikipedia NEs, which come from a different distribution compared to GoodNews, thus degrading the final model performance. 
% As a takeaway, specializing in one domain reduces generalization over non-related ones.

% We provide our exploratory experiments in the lower half of . 
% In this set of experiments, we first inspect the effect of pre-training on Wikipedia's sections and descriptions evaluated on GoodNews~\cite{biten2019good}. We draw two conclusions. Firstly, pre-training on section or description results in better performance in terms of CIDEr and Rouge-L over the baseline model (1st row), both with and without MNEM. 
% Secondly, pre-training on sections is more beneficial than pre-training on descriptions. This is expected since sections are longer and more detailed, having a similar distribution to news articles than descriptions alone. Furthermore, we note that a slight boost in performance is achieved by pre-training with MNEM as a pretext task. 
% % Swayed by these initial experiments, 
% We probe more into the effect of MNEM on GoodNews.
% % with and without transferring the weights from Wikipedia. 
% We notice that MNEM is very effective, % even without any transfer learning, 
% improving the results by 1-2\% overall. 
\subsection{Ablation Study on MNEM}
To further probe into the efficacy of MNEM, we compare it against the well established MLM introduced by BERT. The MLM strategy selects 15\% of the words in a sentence in which it replaces WordPiece tokens~\cite{wu2016google} 80\% of the time with a special [MASK] token, 10\% of the time with a random word and 10\% of the time without changing the word. We introduce another variant of MLM as a comparison namely Full Masking where given 15\% of the words in a sentence, we always mask a word without employing other cases. 
We perform the pre-training stage with these masking strategies on transformer models and report results on Wikipedia Captioning in Table~\ref{table:masking}.
% \textcolor{red}{Table~\ref{table:masking} demonstrates the scores of three distinct masking strategy, divided into 3 various definition of context so that we can examine how context effects the masking strategies.} 

\input{tables/masking}
%First of all, 
We note that our newly defined Full Masking strategy outperforms the well established MLM, challenging the hyper-parameters defined by MLM. Our experiments demonstrate that masking hyper-parameters are rather task dependent which need to be treated selectively. Overall, MNEM yields a significant boost over its counterparts in TaT (\textbf{+12} points in CIDEr), while this behaviour is less apparent in T5 and GPT-2. This is justified since TaT is specifically designed for News Image Captioning, highlighting the effectiveness of MNEM in a fine-grained multi-modal scheme. In contrast, T5 and GPT-2 essentially are language models that have been pretrained on text thus require more careful adaptations in architecture for further improvements.

%second paragraph to highlight the role of each context type
To examine how the context affects the masking strategies, we carry out experiments with TaT, with the context defined in three distinct ways:
%divided into 3 various definitions: 
\textit{description}, \textit{section} and the original \textit{Wiki context}.
We observe that MNEM again outperforms other methods on every context type. Notably, the difference between MNEM and the other two masking strategies is much higher with the description compared to section used as a context. This is somehow expected given that a section tends to be longer and it is harder to generate contextualized captions. However, we observe that MNEM is especially effective when section and description are combined (Wiki as context). MNEM performs \textbf{2\%} better in Recall in Wiki context while improving \textbf{1\%} more in Recall compared to MLM or Full Masking in description or section as a context, this behaviour is more apparent in CIDEr. This demonstrates that MNEM is not only better on encoding the contextual information but especially better at combining description and section information. We believe the potential of MNEM is unlocked further because it can amalgamate the two contexts better by discovering strong correlations between description and section, resulting in better representations for NEs as the numbers clearly suggest.
\subsection{Analysis of Wikipedia Captioning}
In this section, we investigate the performance of our models in Wikipedia Captioning and provide an in-depth analysis in both quantitative and qualitative results. Furthermore, we examine if our models have the ability to generate a caption for ``irrelevant'' images given a context. This evaluates how close our model performs in terms of generating captions for any given image while staying faithful to the section. 
% Our research is intended to identify the key components(copying, visual features guidance) that mainly contribute to its state-of-the-art performance in this particular task. Also, we figure out the limits of Transform-and-Tell in several aspects that it struggles to deal with, as well as illustrate the difference from our proposed method to help the model address its limitation by comparing the network performances.

% \input{tables/desc2cap}
% \input{tables/sec2cap}
% \subsubsection{Desc2Cap}

\input{tables/easy_vs_hard}

\input{figures/story_telling}
\subsubsection{Easy vs Hard Samples.}
Previously, we show in Table~\ref{table:wiki_2_cap} that seq2seq models exploit a shortcut of copying the NEs from the context, reflected in the high precision at inserting NEs. Hence, to better understand our model's limitations, we devise a procedure to identify the captions that can be produced simply by utilizing the context. 
% and disregarding guidance from the visual features. 
To this end, we employ the Jaccard similarity~\cite{jaccard1912distribution} at the word level where given two sets $\mathcal{C}, \mathcal{S}$ for caption and context respectively composed of word tokens, we calculate $J(\mathcal{C}, \mathcal{S}) = \frac{|\mathcal{C} \cap \mathcal{S}|}{|\mathcal{C} \cup \mathcal{S}|}$. 
% The distribution in terms of the Jaccard score between caption and context can be found in supplementary material. 
Accordingly, we assume the image has a diminishing impact on the caption when the caption and the context have a Jaccard score of higher than 0.5. Thus, we refer to these samples as \textit{easy} samples (\textit{hard} otherwise) since the caption can be generated simply by copying from the context and ignoring the image.

% \begin{equation}
%   \begin{cases} 
%       \text{Hard,} & J(\mathcal{C}_i, \mathcal{S}_i) \leq 0.5 \\
%       \text{Easy,} & J(\mathcal{C}_i, \mathcal{S}_i) > 0.5 \\
%   \end{cases}
% \end{equation}
Table~\ref{table:easy_hard_comp} presents the scores in terms of Easy and Hard subset,
%as well as the effect of each type of contextual information separately, when section or description is treated as the whole context. Furthermore, we also supply 
as well as the average Jaccard score between the caption output and the context. 
% Easy vs Hard
%There are several insights to be drawn. 
We observe that the average GT overlap score between Easy and Hard subset (GT. ol.) is quite disparate,
% and this holds true for each type of definition of context. And
yet we do not see a similar distribution learned by our models in generated overlap score (see Gen. ol. in Table~\ref{table:easy_hard_comp}). The models instead have higher overlap scores in both Easy and Hard samples (see Gen. ol.) compared to GT overlap. \textit{This highlights the first limitation of our model where the copying shortcut is exploited and learnt through the Easy subset while being carried over to the Hard samples.}
% This clearly pinpoints the first limitation of our models, in which our models generates captions more similar in nature to Easy subset, disregarding partly the distribution of Hard subset. 
% Metrics
This behaviour is also apparent in the captioning metrics as well as Precision and Recall by the excessive gap between Easy and Hard subset regardless of the models.
% in all the contexts. 
% For example, when description and section is used together as a context (Wikipedia captioning), it can be appreciated that the scores drop from 362 to 160 in CIDEr and 37 to 10 in BLEU-4 for TaT and 367 to 164 in CIDEr and 37 to 11 in BLEU-4 for TaT+MNEM. 
% The same behaviour is also observed in terms of Precision and Recall, significant drop in all the context and 
Nevertheless, the benefits of MNEM are observable, as this method outperforms our base models in both subsets regarding captioning metrics. Improvements in Recall scores, along with increases in length, especially indicate better diversity and NE insertion in generated captions which can be found in all the models. However, MNEM still can not effectively deal with the distribution shift that exists within Wikipedia. In other words, all of the models suffer from relying too much on the contextual information.

% note
% gen Length longer, diversity, increase coverage
% high precision, risk of copying
% Context importance
% Furthermore, given the effect on the Easy subset, we next focus on which context is more influential on this phenomena. According to Table~\ref{table:easy_hard_comp}, both of the models favor copying strategy more when description is used as a context. On the other hand, when section is utilized as context, the models have lower generated overlap score in both subsets. This is somehow expected given the larger amount of information at a fine-grained level in sections. Moreover, captions present a compact version of sections summarizing the whole context instead of just picking out a part of the text to describe the image.

\subsubsection{Context-based Story Telling.} 

Humans have an innate capability of coming up with a well sounding explanation for an image given any context. In order to assess this skill in current captioning pipelines, we employed a retrieval system to sort images according to the degree of relevance given a section. 
Our goal is to evaluate our models' ability to adapt their produced image descriptions based on the context utilized, \textit{i.e.} exploit the given context that comes in the form of Named Entities. 
% As a goal, we would like to have models that can utilize the abundant contextual information to describe the image in different ways, 
% We next discuss this point in-depth in the following section by examining if our models are actually capable of generating diverse captions with unpaired images as input.
In our experiments, we trained two fully-connected layers on top of the visual and textual representations of CLIP~\cite{radford2021learning} to obtain ranked images according to the semantic relatedness in a decreasing order. In Figure~\ref{fig:story}, we provide qualitative samples for the studied models with and without MNEM. We ask the reader to refer to the Appendix for more samples. It can be seen that  models can generate context-based captions even with irrelevant images to a certain degree. For example, even though the images in 1st and 2nd row, 3rd column are irrelevant to the input section, MNEM models can ``hallucinate'' better sounding captions that can improve upon the combination of modalities. 
Also, MNEM models, tend to include more relevant information in the form of Named Entities, thus integrating relevant incoming information from the section.
However, this is not always the case, we can also note that models tend to generate short sentences that are repeated over many images, thus missing some salient details coming from the visual input. All in all, even though improved captions are generated with MNEM, this innate human skill lies far compared to current approaches that favor template-based, short and repetitive sentences, thus opening ways for future work. 
% However. in some cases, they can integrate the visual details into the caption.       

% Training details, as well as performance of this retrieval model in Wikipedia Captioning are provided in the supplementary material. 
% Once the retrieval model was trained, we used a subset of 1K images from the test set of Wikipedia Captioning. 
% By ranking the similarities of the retrieved images according to the sections, we arrange the images according to the semantic relatedness in a decreasing order. We later employ this images as input along with the section to TAT~\cite{tran2020transform} and our proposed model. 
% In Figure TODO we show that....
% Write observations....

\section{Conclusions}
In this work, we have proposed the novel task of Wikipedia Captioning. 
We introduce the importance of the Wikipedia captioning due to the direct impact of it on several applications, as well as advancing visio-linguistic tasks. We show extensive experimentation to compare our approach with several baselines. By incorporating a novel pretext task, namely Masked Named Entity Modelling (MNEM), we model the distribution of NEs as an objective to be optimized. We show that by integrating this approach, we reach top performance on Wikipedia captioning. 
We also show that a model pre-trained on Wikipedia and optimized with MNEM generalizes well to the News Captioning task. 
% \section{Limitations and Broad Impact}
% Despite the encouraging results, 
We found that baselines tend to learn a shortcut to the captioning task and directly copy the image description to generate a caption. To properly evaluate this effect, we divide the Wikipedia Captioning test set according to the Jaccard similarity between the context and ground-truth captions. We show that there is still a big gap among these two test splits, which opens the path for future research.
Finally, we show qualitative examples regarding the human capability of generating captions adapted to any context.
% even when there is no explicit pairing among them. 
We show limitations that current models have and provide insights of the benefits of our approach.
% As future work, we would like to incorporate information coming from retrieval relevance into an end-to-end model that is capable of suggesting relevant images and captions given a context. Such a model could benefit of optimizing both tasks to obtain improvements in the task proposed in this work.

\bibliography{main}

\clearpage
\appendix
\section{Technical Appendix}
\label{sec:appendix}

\subsection{Dataset}
The WIT\footnote{\url{https://github.com/google-research-datasets/wit}} dataset contains over 37.6 million data points based on a Wikipedia Crawl. Originally, this dataset is proposed for the image-text retrieval task with special emphasis on multilingual text. 
% We perform experiments on WIT training data obtained from the official repository\footnote{\url{https://github.com/google-research-datasets/wit}} which contains over 37.6 million image-text sets. 
% To make it fit to the training pipelines, 
We discard all the data that comes from non-English Wikipedia pages and get a subset of around 5.4M samples. In addition, we perform extra pre-processing steps by removing short text, rare punctuation and also filter all images in uncommon format and resize into 256x256. 
For each data point including an image, we take the \textit{caption\_reference\_description} as the ground truth caption, \textit{caption\_attribution\_description} as the description and \textit{context\_section\_description} as the associated section. Depending on the context, we accordingly de-duplicate the dataset to assure all the triplets (image, context, caption) unique. All the filter steps reduce the training data to a final dataset of total over 2.6M samples. 
% with around 2M images. 
We create a split by randomly partitioning the training data to 2.6M training, 8k validation, 20K for testing, as the official split is not publicly available at the time of this work. 

For News Image Captioning experiments, we train the model on the GoodNews dataset using the 421K training, 18K validation, and 23K test split, provided by~\cite{biten2019good}.
Similar to WIT, each data sample in this dataset is a triplet consisting of an article as the whole context, an image, and a caption.
% Table shows some statistics of the split we use. e. For more details of the dataset used, we ask the reader to refer to the supplementary material section.
% \subsection{Context-Based Story Telling}
% This is a section in the appendix.
% Data statistics
\subsection{Baselines} 
We define several baselines, which are composed of an LSTM encoder-decoder, to serve as comparison to our proposed method. 
Particularly, we frame Wikipedia captioning as a machine translation problem in which we treat (image, section, description) sequence as a source language and the caption as a target language. We thus consider machine translation sequence-to-sequence (Seq2seq) model and its incremental variants. For this experimentation, input image features extracted from a ResNet-152 are reused whereas a SentencePiece tokenizer~\cite{kudo-richardson-2018-sentencepiece} is trained for text tokenization.

\noindent
\textbf{Seq2seq.} The source of image captioning models~\cite{vinyals2015show,Xu2015,anderson2017bottom} can be associated to the success of machine translation sequence-to-sequence (seq2seq) models~\cite{Bahdanau2014,Sutskever}. Thus, for a robust baseline we treat the Wikipedia captioning as a machine translation task in which we treat section with description as a source language and the caption as a target language. 

\noindent
\textbf{Seq2seq with Attention.} The next advance in image captioning was the introduction of attention module~\cite{Xu2015} which was also inspired from machine translation field~\cite{Bahdanau2014}. We use the attention block in the decoder to attend to the encoded features at each time step. 

\noindent
\textbf{Seq2seq with Pointer.} The introduction of pointer mechanism~\cite{vinyals2015pointer} allowed the seq2seq models to have a supervisory signal on attention weights. It was mostly used for out-of-dictionary words to directly copy it from the encoder. 

\noindent
\textbf{Seq2seq with Pointer and Coverage.} The coverage idea was introduced by~\cite{see2017get} to overcome the repetition problem that seq2seq models suffering. Conceptually, they penalize the attention weights that repeatedly attends to the same locations.

Experiments quantitatively show the robustness of the baselines in this task, despite of its simplicity, with competitive performances compared to large transformer counterparts including state-of-the-art methods in News Image Captioning. This not only explains the baseline choice, but also reveals one of the most challenging aspects of Wikipedia Captioning: a \textbf{strong linguistic bias} intrinsically embedded in the Wikipedia data, misleading a model to adversely generalize over a specific dataset subset. Hence, a mechanism for effectively reasoning over multi-modal Wikipedia input is essential to escape this sub-optimal convergence.

\subsection{Implementation Details}
We base our implementation on the code provided by TaT\footnote{\url{https://github.com/alasdairtran/transform-and-tell}} and Huggingface Transformers~\cite{wolf-etal-2020-transformers}. 

\noindent
\textbf{Transformer models} We use the same configuration in terms of architecture as proposed in its original papers. We utilize the pre-trained tokenizer available in each model to decompose text inputs. In our pre-training stage, we use the Name Entity Recognition (NER) model available from SpaCy~\cite{spacy2} library. We mask 80\% of NEs at token level with [MASK] token which is added to the model's built-in vocabulary if not included. From NER output, we only mask NEs with labels
of: [PERSON, ORG, GPE] for Wikipedia and additional ones of [NORP, LOC, FAC] in case of GoodNews. In all experiments, we use a batch size of 16, and train the model for 3 epochs on WIT, and 8 epochs on GoodNews.
For parameter optimizations, 
% We use AdamW~\cite{loshchilov2017decoupled} optimizer with the learning rate linearly warmed up to $10^{−4}$ in the first 5 epochs and decay with the rate of $10^{−4}$ afterwards. In the fine-tuning phase, we set the initial learning rate to $10^{−5}$ with the same weight decay and train the model for 100 epochs if an early stop is not achieved. We define the early stop strategy as no improvement on validation set after 10 consecutive epochs. 
we use AdamW~\cite{loshchilov2017decoupled} with $\beta^1=0.9$ and $\beta^2=0.98$, both in pre-training and fine-tuning. In pre-training stage, we use learning rate of $10^{-4}$ linearly warmed up during the first 5\% training steps, while at fine-tuning, we set the initial learning rate to $10^{-5}$ with the same weight schedule.
We use COCO evaluation toolkit~\cite{lin2014microsoft} for the evaluation of our models.
\noindent
\textbf{Baselines} Models are trained with the OpenNMT~\cite{klein2017opennmt} library, in which all of our models are composed of an LSTM encoder-decoder. In all of our models we follow the same settings. We utilize batch size of 32 with Adam~\cite{kingma2014adam} optimizer, accumulating the gradients for 3 iterations before forwarding the loss. We set the learning rate to $5\times 10^{-4}$ and employ gradient clipping set to 5. We use a bi-directional LSTM for encoder and a LSTM for decoder with hidden and embedding size is set to 512 and employ dropout with 0.1 probability. All of our models are trained for 100K iterations with inputs tokenized with SentencePiece tokenizer.
\input{tables/nic_models}
\subsection{WiT Statistics}
% \label{app:wit_stats}
We supply statistics on the WIT data split mentioned in the main paper. As can be appreciated from Table~\ref{tab:wit_stats}, around 34\% of the words in the captions are Named Entities (NEs) and description follows a very similar distribution in terms of NEs as well. 
%Moreover, we observe that section contains less NEs compared to caption and description. 
We also provide the average length of each domain. As observed, captions are shorter in length which explains the behaviour of our models generating short sentences. However, we do see that the average length of sections is significantly higher compared to descriptions.
\input{tables/wit_stats}

% Distribution of Test Set
As mentioned in the main paper, we divide the test set into two subsets, namely Easy and Hard. These subsets are defined according to Jaccard similarity (\textit{i.e.} Intersection over Union). This setting discloses a shortcut in which our models exploit plain text similarities existing in the Easy set. Here, we present a distribution of the Jaccard similarity according to three types of context in Figure~\ref{fig:jacc_dist}. Clearly, description has a higher overlap with captions than sections, and its combination even provides stronger correlation. Hence, it is expected that models exploit a copying mechanism directly from the context, especially descriptions rather than sections. Furthermore, we see that there are more samples that have Jaccard similarity higher than 0.5, leading the models to ``overfit'' to the Easy set distribution instead of focusing on the Hard set.
\input{figures/jaccard_dist}
\subsection{WIT vs GoodNews}
We provide a comparison between WIT and GoodNews in Table~\ref{tab:wit_vs_gns}. The first advantage of Wikipedia over News captioning is the amount of data available to train our models. Also, Wikipedia has a lower caption length compared to GoodNews (8.88 to 18.21). This could prompt us to interpret Wikipedia Captioning as a simpler task in terms of grammar and generation-wise correctness compared to GoodNews. Nevertheless, we observe that in terms of NEs, Wikipedia contains a higher percentage of NEs at word level (35\% in Wiki to 19.5\% in GoodNews). Hence, in Wikipedia captioning, the difficulty lies in selecting, injecting or generating more relevant NEs on average than News captioning while another difficulty is the highly contextualised nature of the task (same image used in different articles).

% Distribution of NEs
Moreover, Figure 4 shows how different NE categories are distributed within both datasets. Aside from the magnitude, we observe a similarity between these two distributions which motivates us to use Wikipedia as a probing domain for News captioning. Among all, people’s names (PERSON), geopolitical entities (GPE) and organizations (ORG) are the most prevalent NE categories. As a result, we perform MNEM pre-training focusing on these NEs while regarding to the rest as context.
\input{figures/wit_v_gns}
\input{tables/wit_vs_gns}

\subsection{More on effects of different types of context}
% Context importance
Given the effect of the Easy subset on the copy behavior as figured out in the main paper, we investigate the influence of three forms of context on this phenomenon: \textit{description}, \textit{section} and \textit{Wiki}. We conduct experiments on TaT with MNEM as this model shows to be effective in dealing with multi-modal inputs. 
Table~\ref{table:easy_v_hard_context} presents the results corresponding to each type of contextual information. When description is used as context, both models favor copying strategy (see Gen. ol.), demonstrating that description has a greater impact on this behavior. On the other hand, when section is utilized as context, the models have lower generated overlap score in both subsets. This is somehow expected given the larger amount of information at a fine-grained level in sections. Moreover, captions present a compact version of sections summarizing the whole context instead of just picking out a part of the text to describe the image.
\input{tables/easy_vs_hard_context}

\subsection{Methods in News Image Captioning}
Table~\ref{table:nic_models} provides a high-level summary of transformer models for News Captioning. Overall, these methods still rely on prominent pre-trained encoders to independently process multi-modal inputs while leave the burden of both reasoning and generating on a decoder.
%the need of having a effective mechanism for rich-context domain
%with/without extra step for NEs 
However, our proposed method aims to explore the potential of rich semantic and linguistic knowledge in PLMs, effectively freeing the model from the dependency on the external text encoder and templates for handling NEs. In addition, our pre-training strategy in Wikipedia Captioning sufficiently equips the model with Wikipedia knowledge, which is useful to perform similar tasks in different domains.
\input{tables/goodnews_ext}

\subsection{More on GoodNews}
We conduct additional experiments on GoodNews with results reported in Table~\ref{table:goodnews_ext}. As observed, MNEM pre-training improves the performance on both TaT and T5, showing the efficacy of NE-centric strategy in News domain. On the other hand, there is a big gap in performance between $\text{T5}_{\textit{BASE}}$ compared to $\text{T5}_{\textit{BASE}}\texttt{++}$ (see Table~\ref{table:goodnews} in the main paper), suggesting that our model can effectively reason on multi-modal inputs rather than text-only inputs. 

%\subsection{MNEM pre-training on T5}
% \begin{figure}[t!]
% \centering
%     \includegraphics[width=\columnwidth, height=0.1\textheight]{figures/pdfs/MNEM-T5.pdf}
% \caption{\textbf{Overview of MNEM pre-training task designed for T5}}
% \label{fig:mnem_t5}
% \end{figure}

\subsection{Qualitative Samples on Wikipedia Captioning}
In this section, we show qualitative samples on the Wikipedia captioning to showcase our models capabilities. 
As can be seen, all of the models tend to directly copy descriptions into the generated captions. Moreover, we show that T5 and TaT pre-trained with MNEM usually generate NEs with more variety and higher precision. 
\input{figures/wiki_qual_results}

\end{document}

%% file: tables/wiki_cap.tex
\setlength{\tabcolsep}{4pt}
\begin{table}[t!]
\tiny
\begin{center}
\caption{\textbf{Performance of proposed models on Wikipedia Captioning.} Seq2seq performs better than TaT without image while MNEM is effective to align the image and text modality. Metric scores are reported as percentage. $\dagger$ indicates that MNEM pre-training is performed. $\Delta$ indicates the improvement with MNEM compared to respective models. Bold numbers are the best performance of each metric.}
\label{table:wiki_2_cap}
\begin{adjustbox}{width=1\columnwidth}
\begin{tabular}{clccccccc}
\hline\noalign{\smallskip}
& Model & B & M & R & C & S & Pr & Re\\
\noalign{\smallskip}
\hline
\noalign{\smallskip}
% \parbox[t]{2mm}{\multirow{5}{*}{\rotatebox[origin=c]{90}{No Image}}}& & & & & & & & \\
\parbox[t]{2mm}{\multirow{13}{*}{\rotatebox[origin=c]{90}{No Image}}} & Seq2seq & 6.04 & 11.53 & 31.94 & 137.38 & 20.25 & 26.42 & 14.05\\
& Seq2seq+Attn. & 15.86 & 18.82 & 43.69 & 242.87 & 31.45 & \textbf{44.10} & 27.60\\		
& Seq2seq+Ptr. & 21.04 & 21.27 & 45.22 & 255.87 & 32.64 & 40.81 & 30.67 \\	
& Seq2seq+Ptr.+Cvrg. & 21.57 & 21.56 & 45.50 & 258.84 & 33.00 & 41.23 & 31.37\\	
& $\text{TaT}$ & 18.74 & 19.65 & 43.74 & 229.77 & 29.47 & 32.09 & 27.12\\
& $\text{GPT-2}_{\textit{BASE}}$ & 19.18 & 20.68 & 45.68 & 254.94 & 32.8 & 42.28 & 30.38 \\
& $\text{T5}_{\textit{BASE}}$ & \textbf{25.70} & \textbf{24.17} & 48.8 & 275.31 & 34.31 & 39.11 & \textbf{35.09}\\
\noalign{\smallskip}
\cline{2-9}
\noalign{\smallskip}
& $\text{TaT}^\dagger$ & 19.93 & 20.42 & 44.57 & 235.9 & 30.03 & 32.94 & 28.47\\
& $\text{GPT-2}_{\textit{BASE}}^\dagger$ & 19.64 & 21 & 46.32 & 260.4 & 33.34 & 43.31 & 30.9\\
& $\text{T5}_{\textit{BASE}}^\dagger$ & 24.60 & 23.65 & \textbf{48.96} & \textbf{278.68} & \textbf{34.69} & 40.51 & 34.26\\
\noalign{\smallskip}
\cline{2-9}
\noalign{\smallskip}
& $\Delta$ $\text{TaT}$ & 1.19$\uparrow$ & 0.77$\uparrow$ & 0.83$\uparrow$ & 6.13$\uparrow$ & 0.56$\uparrow$ & 0.85$\uparrow$ & 1.35$\uparrow$\\
& $\Delta$ $\text{GPT-2}_{\textit{BASE}}$ & 0.46$\uparrow$ & 0.32$\uparrow$ & 0.64$\uparrow$ & 5.46$\uparrow$ & 0.54$\uparrow$ & 1.03$\uparrow$ & 0.52$\uparrow$\\
& $\Delta$ $\text{T5}_{\textit{BASE}}$ & 1.1$\downarrow$ & 0.52$\downarrow$ & 0.16$\uparrow$ & 2.08$\uparrow$ & 0.38$\uparrow$ & 1.4$\uparrow$ & 0.83$\downarrow$\\
\noalign{\smallskip}
\hline
\noalign{\smallskip}
\parbox[t]{2mm}{\multirow{13}{*}{\rotatebox[origin=c]{90}{Image}}} & Seq2seq & 6.10 & 11.26 & 31.20 & 136.98 & 19.83 & 25.97 & 13.88\\
& Seq2seq+Attn. & 14.71 & 18.06 & 42.79 & 235.44 & 30.70 & \textbf{43.49} & 26.60\\
& Seq2seq+Ptr.& 20.46 & 20.89 & 44.76 & 252.79 & 32.12 & 40.37 & 29.97\\
& Seq2seq+Ptr.+Cvrg. & 20.22 & 20.72 & 44.45 & 249.67 & 31.77 & 40.20 & 29.68\\
& $\text{TaT}$ & 24.47 & 23.15 & 48.07 & 267.37 & 33.33 & 32.10 & 34.64\\
% \hline
& $\text{GPT-2}_{\textit{BASE}}\texttt{++}$& 19.4 & 20.72 & 45.97 & 256.25 & 32.79 & 41.28 & 30.06\\
& $\text{T5}_{\textit{BASE}}\texttt{++}$ & 23.83 & 23.33 & 48.8 & 276.6 & 34.59 & 40.7 & 33.72\\
\noalign{\smallskip}
\cline{2-9}
\noalign{\smallskip}
& $\text{TaT}^\dagger$ & 25.31 & 23.64 & 48.72 & 272.46 & 34.36 & 33.34 & 35.56\\
& $\text{GPT-2}_{\textit{BASE}}^\dagger\texttt{++}$ & 20.65 & 21.54 & 47.3 & 266.29 & 34.12 & 42.86 & 31.48\\
& $\text{T5}_{\textit{BASE}}^\dagger\texttt{++}$ & \textbf{25.89} & \textbf{24.39} & \textbf{49.28} & \textbf{279.09} & \textbf{34.64} & 39.65 & \textbf{35.73}\\
\noalign{\smallskip}
\cline{2-9}
\noalign{\smallskip}
& $\Delta$ $\text{TaT}$ & 0.84$\uparrow$ & 0.49$\uparrow$ & 0.65$\uparrow$ & 5.09$\uparrow$ & 0.73$\uparrow$ & 1.24$\uparrow$ & 0.92$\uparrow$\\
& $\Delta$ $\text{GPT-2}_{\textit{BASE}}\texttt{++}$ & 1.25$\uparrow$ & 0.82$\uparrow$ & 1.33$\uparrow$ & 10.04$\uparrow$ & 1.33$\uparrow$ & 1.58$\uparrow$ & 1.12$\uparrow$\\
& $\Delta$ $\text{T5}_{\textit{BASE}}\texttt{++}$ & 2.56$\uparrow$ & 1.06$\uparrow$ & 0.48$\uparrow$ & 3.78$\uparrow$ & 0.05$\uparrow$ & 1.05$\downarrow$ & 2.01$\uparrow$\\
\noalign{\smallskip}
\hline
\end{tabular}
\end{adjustbox}
\end{center}
\vspace{-.8cm}
\end{table}
\setlength{\tabcolsep}{1.4pt}

%% file: tables/goodnews.tex
\setlength{\tabcolsep}{4pt}
\begin{table}[t!]
\tiny
\begin{center}
\caption{\textbf{Performance of proposed models on GoodNews.} Ft means fine-tuning the models with the captioning objective in the corresponding dataset. Note that for captioning metrics, we directly use the results that are reported in the original papers.}
\label{table:goodnews}
\begin{adjustbox}{width=1\columnwidth}
\begin{tabular}{c|cc|cc|cccccc}
\hline\noalign{\smallskip}
& \multicolumn{2}{c}{WIT} & \multicolumn{2}{c|}{GoodNews} & \multirow{2}{*}{B} & \multirow{2}{*}{M} & \multirow{2}{*}{R} & \multirow{2}{*}{C} & \multirow{2}{*}{Pr} & \multirow{2}{*}{Re}\\
\noalign{\smallskip}
\cline{2-5}
\noalign{\smallskip}
& MNEM & Ft & MNEM & Ft &  &  &  &  &  & \\
\noalign{\smallskip}
\hline
\noalign{\smallskip}	
$\text{TaT}$ &  &  &  & $\checkmark$ & 6 & 10.1& 21.2  & 53.1 & 21.3 & 19.7\\
$\text{TaT(Full)}$ &  &  &  & $\checkmark$ & 6.05 & 10.3 & 21.4 & 53.8 & 21.85 & 19.93\\
$\text{VisualNews(Full)}$ &  &  &  & $\checkmark$ & 6.1 & 8.3 & 21.6 & 55.4 & -- & --\\
$\text{JoGANIC}$ &  &  &  & $\checkmark$ & 6.34 & 10.78 & 21.65 & 59.19 & -- & --\\
$\text{JoGANIC(Full)}$ &  &  &  & $\checkmark$ & 6.83 & \textbf{11.25} & \textbf{23.05} & 61.22 & -- & --\\
\noalign{\smallskip}
\hline
\noalign{\smallskip}						
\multirow{8}{*}{$\text{T5}_{\textit{BASE}}\texttt{++}$} &  &  &  & $\checkmark$  & 6.34 & 10.05 & 21.8 & 59.68 & 23.61 & 20.08\\
&  &  & $\checkmark$ & $\checkmark$ & 6.56 & 10.21 & 21.71 & 60.17 & 23.28 & 20.43\\
&  & $\checkmark$ &  & $\checkmark$ & 7.17 & 10.8 & 22.6 & 62.76 & 23.11 & 20.92\\
&  & $\checkmark$ & $\checkmark$ & $\checkmark$  & \textbf{7.19} & 10.84 & 22.89 & \textbf{63.89} & \textbf{23.37} & \textbf{21.48}\\
& $\checkmark$ &  &  & $\checkmark$ & 5.89 & 9.38 & 19.88 & 51.49 & 20.36 & 18.85\\
& $\checkmark$ &  & $\checkmark$ & $\checkmark$ & 6.63 & 10.19 & 21.16 & 57.03 & 21.51 & 20.43\\
& $\checkmark$ & $\checkmark$ &  & $\checkmark$ & 6.66 & 10.28 & 21.41 & 58.47 & 21.84 & 20.59\\
& $\checkmark$ & $\checkmark$ & $\checkmark$ & $\checkmark$  & 6.75 & 10.38 & 21.53 & 59.07 & 21.89 & 20.55\\
\noalign{\smallskip}
\hline
\end{tabular}
\end{adjustbox}
\end{center}
\vspace{-.7cm}
\end{table}
\setlength{\tabcolsep}{1.4pt}

%% file: tables/masking.tex
\setlength{\tabcolsep}{4pt}
\begin{table}[t!]
\tiny
\begin{center}
\caption{\textbf{Ablation studies on different masking strategies.} Our proposed method MNEM outperforms MLM and Full masking strategy on every context and on every metrics. All the numbers are obtained in Wiki test set.}
\label{table:masking}
\begin{adjustbox}{width=1\columnwidth}
\begin{tabular}{cccccccccc}
\hline\noalign{\smallskip}
Model & Context & Masking & B & M & R & C & S & Pr. & Re.\\
\noalign{\smallskip}
\hline
\multirow{3}{*}{$\text{GPT-2}_{\textit{BASE}}\texttt{++}$} & \parbox[t]{2mm}{\multirow{3}{*}{Wiki.}} & Full & 20.44 & 21.47 & 47.3 & \textbf{266.4} & 34.06 & 43.24 & 31.4\\
 & & MLM & 20.34 & 21.38 & 47.3 & 266. & 34.03 & \textbf{43.4} & 31.4\\
 & & MNEM & \textbf{20.65} & \textbf{21.54} & \textbf{47.3} & 266.29 &\textbf{34.12} & 43.12 &\textbf{31.55}\\
\hline
\multirow{3}{*}{$\text{T5}_{\textit{BASE}}\texttt{++}$} & \parbox[t]{2mm}{\multirow{3}{*}{Wiki.}} & Full & 2502 & 24 & 49.04 & 276.58 & 34.45 & 38.87 & 35.08 \\
 & & MLM & 24.57 & 2371 & 48.88 & 275.22 & 34.37 & 38.75 & 34.94\\
 & & MNEM & \textbf{25.89} & \textbf{24.39} & \textbf{49.28} & \textbf{279.09} &\textbf{34.64} &\textbf{39.65} &\textbf{35.73}\\
\hline
\multirow{9}{*}{$\text{TaT}$} &
\parbox[t]{2mm}{\multirow{3}{*}{Wiki.}} & Full & 24.01 & 22.84 & 47.49 & 260.87 & 32.79 & 31.83 & 34.35\\
 & & MLM & 23.81 & 22.74 & 47.21 & 258.25 & 32.65 & 31.47 & 34.12\\
 & & MNEM & \textbf{25.31} & \textbf{23.64} & \textbf{48.72} &\textbf{272.46} & \textbf{34.06} & \textbf{33.34} & \textbf{35.56}\\
\cline{2-10} & 
\parbox[t]{2mm}{\multirow{3}{*}{Sec.}} & Full & 8.96 & 11.98 & 32.13 & 147.43 & 19.71 & 20.02 & 15.13\\
 & & MLM & 8.71 & 11.40 & 32.07 & 147.40 & 19.59 & 19.81 & 14.46\\
 & & MNEM & \textbf{9.54} & \textbf{12.52} & \textbf{33.03} & \textbf{153.87} & \textbf{20.37} & \textbf{20.75} & \textbf{16.01}\\
\cline{2-10} &
\parbox[t]{2mm}{\multirow{3}{*}{Desc.}} & Full & 20.95 & 20.39 & 42.10 & 215.42 & 27.97 & 28.80 & 26.55\\
 & & MLM & 20.92 & 20.31 & 41.86 & 213.69 & 27.73 & 28.41 & 26.16\\
 & & MNEM & \textbf{22.36} & \textbf{21.29} & \textbf{43.14} & \textbf{224.38} &\textbf{ 28.87} &\textbf{29.36} &\textbf{27.90}\\
\hline
\end{tabular}
\end{adjustbox}
\end{center}
\vspace{-.7cm}
\end{table}
\setlength{\tabcolsep}{1.4pt}

%% file: tables/easy_vs_hard.tex
\setlength{\tabcolsep}{4pt}
\begin{table}[t!]
\tiny
\begin{center}
\caption{\textbf{Performance on Easy and Hard subsets.} They are defined based on the Jaccard similarity between the context and caption. Overlap scores measure the similarity compared to the context, reported in average Jaccard score. We refer the (ground truth caption, context) overlap as GT ol., Gen. ol. is the (generated caption, context) overlap and Gen. len. is the average length of generated captions. $\dagger$ indicates that MNEM pre-training is performed.}
\label{table:easy_hard_comp}
\begin{adjustbox}{width=1\columnwidth}
\begin{tabular}{lrccccccc}
\hline\noalign{\smallskip}
Model & Test Set & GT ol. & Gen. ol. & Gen. len. & B & C & Pr & Re\\
\hline
\parbox[t]{2mm}{\multirow{3}{*}{$\text{TaT}$}} & Easy & 0.81 & 0.92 & 7.65 & 29.93 & 313.99 & 39.49 & 37.2\\
& Hard & 0.35 & 0.75 & 6.03 & 8.48 & 133.68 & 19.06 & 16.06\\
& Full & 0.71 & 0.88 & 6.46 & 24.47 & 267.37 & 32.10 & 34.64 \\
\hline
\parbox[t]{2mm}{\multirow{3}{*}{$\text{TaT}^\dagger$}} & Easy & 0.81 & 0.92 & 7.89 & 30.76 & 319.01 & 39.71 & 38.20\\
 & Hard & 0.35 & 0.73 & 6.16 & 8.91 & 137.81 & 19.39 & 16.53\\
 & Full & 0.71 & 0.87 & 7.08 & \textbf{25.31} & \textbf{272.46} & \textbf{33.34} & \textbf{35.56}\\
\hline
\parbox[t]{2mm}{\multirow{3}{*}{$\text{GPT-2}_{\textit{B}}\texttt{++}$}} 
 & Easy & 0.81 & 0.93 & 5.35 & 24.74 & 304.71 & 46.37 & 34.24\\
 & Hard & 0.35 & 0.81 & 4.2 & 6.38 & 128.15 & 24.28 & 15.84\\
 & Full & 0.71 & 0.90 & 5.06 & 19.4 & 256.25 & 41.28 & 30.36\\
\hline 
\parbox[t]{2mm}{\multirow{3}{*}{$\text{GPT-2}_{\textit{B}}^\dagger\texttt{++}$}} 
 & Easy & 0.81 & 0.93 & 5.49 & 25.63 & 312.16 & 47.69 & 35.87\\
 & Hard & 0.35 & 0.8 & 4.38 & 7.01 & 133.64 & 24.97 & 16.92\\
 & Full & 0.71 & 0.89 & 5.2 & \textbf{20.65} & \textbf{266.29} & \textbf{42.86} & \textbf{31.48}\\
\hline
\parbox[t]{2mm}{\multirow{3}{*}{$\text{T5}_{\textit{B}}\texttt{++}$}} 
 & Easy & 0.81 & 0.93 & 6.35 & 29.55 & 325.32 & 45.37 & 38.37\\
 & Hard & 0.35 & 0.8 & 4.92 & 7.46 & 130.79 & 23.7 & 18.29\\
 & Full & 0.71 & 0.9 & 5.98 & 23.83 & 276.6 & \textbf{40.7} & 33.72\\
\hline
\parbox[t]{2mm}{\multirow{3}{*}{$\text{T5}_{\textit{B}}^\dagger\texttt{++}$}}
 & Easy & 0.81 & 0.91 & 7.16 & 31.96 & 330.79 & 43.9 & 40.9\\
 & Hard & 0.35 & 0.79 & 5.39 & 7.95 & 129.84 & 22.63 & 18.57\\
 & Full & 0.71 & 0.88 & 6.7 & \textbf{25.89} & \textbf{279.09} & 39.65 & \textbf{35.73}\\
\hline
\end{tabular}
\end{adjustbox}
\end{center}
\vspace{-0.7cm}
\end{table}
\setlength{\tabcolsep}{1.4pt}

%% file: figures/story_telling.tex
\begin{figure*}[t!]
    \centering
    \scriptsize
    \setlength{\tabcolsep}{0em}
    \begin{tabular}{c}
    \toprule
    
  \begin{tikzpicture}

  \tikzset{block/.append style={text width=47em, minimum height=2em, execute at begin node=\tiny,  align=justify}}
    \node[block]                 (section) {Paul Allman Siple December 18 1908 – November 25 1968 was an American Antarctic explorer and geographer who took part in six Antarctic expeditions including the two Byrd expeditions of 1928–1930 and 1933–1935 representing the Boy Scouts of America as ...
    % In addition to being an Eagle Scout Siple was also a Sea Scout His first and third books covered these adventures With Charles F. Passel he developed the wind chill factor and Siple coined the term (...)
    };
    \node[inner sep=0pt, above left=-0.75cm and 0.05cm of section] (logo) 
    {\includegraphics[width=.05\textwidth]{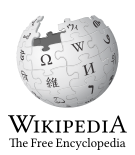}};
    
    \node[inner sep=0pt, below left=0.51em and -7.5em of section] (img1)     {\includegraphics[width=.24\textwidth, height=.13\textwidth]{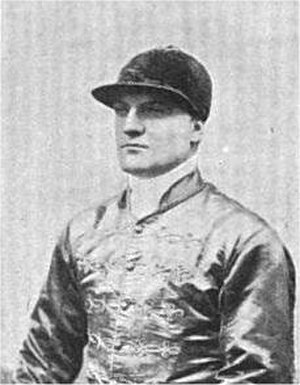}};
    \node[inner sep=0pt, right=0.5em of img1] (img2)     {\includegraphics[width=.24\textwidth, height=.13\textwidth]{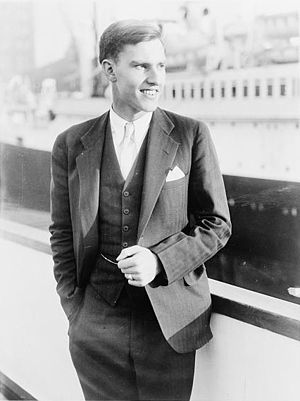}};
    \node[inner sep=0pt, right=0.5em of img2] (img3)     {\includegraphics[width=.24\textwidth, height=.13\textwidth]{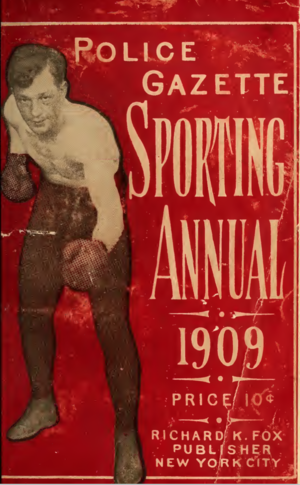}};
    \node[inner sep=0pt, right=0.5em of img3] (img4)     {\includegraphics[width=.24\textwidth, height=.13\textwidth]{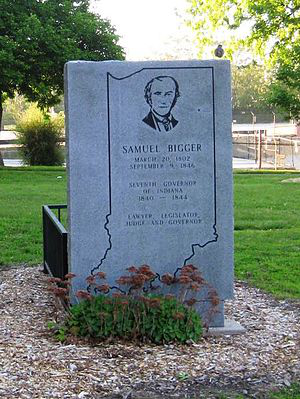}};
    
    \node[signal, draw,minimum width=10cm, signal from=west, signal to=east,shade, right color=red!25, left color=green!25, below=0.2em of section] (rarrow) {\tiny{ranked list of retrieved images}};
    
    %  \node[inner sep=0pt, text width=.24\textwidth,execute at begin node=\tiny, below=0.5em of img1] (tat1) {\color{blue}Paul Allman Siple};
    \node[inner sep=0pt, text width=.24\textwidth,execute at begin node=\tiny, below=0.5em of img1] (mnem1) {\color{teal}Siple in 1930};
    \node[inner sep=0pt, text width=.24\textwidth,execute at begin node=\tiny, below=0.5em of mnem1] (mnem1_2) {\color{orange}Paul Allman Siple, c. 1930};
    \node[inner sep=0pt, text width=.24\textwidth,execute at begin node=\tiny, below=0.5em of mnem1_2] (mnem1_3) {\color{violet}Paul Allman Siple};

    % \node[inner sep=0pt, text width=.24\textwidth,execute at begin node=\tiny, below=0.5em of img2] (tat2) {\color{blue}Siple in 1930};
    \node[inner sep=0pt, text width=.24\textwidth,execute at begin node=\tiny, below=0.5em of img2] (mnem2_1) {\color{teal}Siple in 1930};
    \node[inner sep=0pt, text width=.24\textwidth,execute at begin node=\tiny, below=0.5em of mnem2_1] (mnem2_2) {\color{orange}Paul Allman Siple on the Byrd Antarctic};
    \node[inner sep=0pt, text width=.24\textwidth,execute at begin node=\tiny, below=0.5em of mnem2_2] (mnem2_3) {\color{violet}Paul Allman Siple};    
    
    % \node[inner sep=0pt, text width=.24\textwidth,execute at begin node=\tiny, below=0.5em of img3] (tat3) {\color{blue}Siple in the Antarctic Peninsula};
    \node[inner sep=0pt, text width=.24\textwidth,execute at begin node=\tiny, below=0.5em of img3] (mnem3_1) {\color{teal}Siple in the 1930s};
    \node[inner sep=0pt, text width=.24\textwidth,execute at begin node=\tiny, below=0.5em of mnem3_1] (mnem3_2) {\color{orange}Siple's Eagle Scout badge};
    \node[inner sep=0pt, text width=.24\textwidth,execute at begin node=\tiny, below=0.5em of mnem3_2] (mnem3_3) {\color{violet}Cover of Siple's first book};
    
    % \node[inner sep=0pt, text width=.24\textwidth,execute at begin node=\tiny, below=0.5em of img4] (tat4) {\color{blue}Siple 's grave at the National Park};
    \node[inner sep=0pt, text width=.24\textwidth,execute at begin node=\tiny, below=0.5em of img4] (mnem4_1) {\color{teal}Siple 's grave at the National Museum of the United States};
    \node[inner sep=0pt, text width=.24\textwidth,execute at begin node=\tiny, below=0.5em of mnem4_1] (mnem4_2) {\color{orange}Siple's signature};
    \node[inner sep=0pt, text width=.24\textwidth,execute at begin node=\tiny, below=0.5em of mnem4_2] (mnem4_3) {\color{violet}Grave at Arlington National Cemetery};
  \end{tikzpicture}
  \\
  \midrule
%   \\

  \begin{tikzpicture}
  \tikzset{block/.append style={text width=47em, minimum height=2em,execute at begin node=\tiny,  align=justify}}
    \node[block]                 (section) {The Bela Vista Park Portuguese Parque da Bela Vista is one of the largest open areas within the city limits of Lisbon. Portugal Bela Vista Beautiful View is the name of its adjacent neighborhood. The park comprises an area of 85,000 m² and is often used as a venue for large concerts ...
    % including several Rock in Rio events Many artists such as The Rolling Stones Paul McCartney Rammstein Ben Harper Britney Spears Alicia Keys Shakira Red Hot Chili Peppers Sting Amy Winehouse Lenny Kravitz Anastacia Bon Jovi Linkin Park Guns N Roses Muse Miley Cyrus MIKA Queen Fergie (...)
    };
    \node[inner sep=0pt, above left=-0.85cm and 0.05cm of section] (logo) 
    {\includegraphics[width=.05\textwidth]{figures/qual_samples/Wikipedia-logo.png}};
    
    \node[inner sep=0pt, below left=0.51em and -7.5em of section] (img1)     {\includegraphics[width=.24\textwidth, height=.13\textwidth]{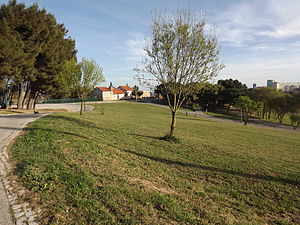}};
    \node[inner sep=0pt, right=0.5em of img1] (img2)     {\includegraphics[width=.24\textwidth, height=.13\textwidth]{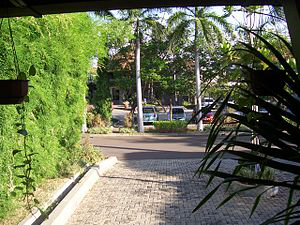}};
    \node[inner sep=0pt, right=0.5em of img2] (img3)     {\includegraphics[width=.24\textwidth, height=.13\textwidth]{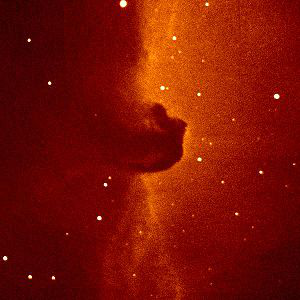}};
    \node[inner sep=0pt, right=0.5em of img3] (img4)     {\includegraphics[width=.24\textwidth, height=.13\textwidth]{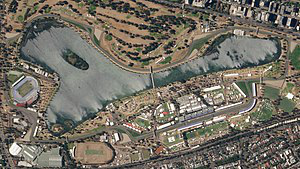}};
    
    \node[signal, draw,minimum width=10cm, signal from=west, signal to=east,shade, right color=red!25, left color=green!25, below=0.2em of section] (rarrow) {\tiny{ranked list of retrieved images}};
    
    % \node[inner sep=0pt, text width=.24\textwidth,execute at begin node=\tiny, below=0.5em of img1] (tat1) {\color{blue}The park in winter};
    \node[inner sep=0pt, text width=.24\textwidth,execute at begin node=\tiny, below=0.5em of img1] (mnem1_1) {\color{teal}The park in winter};
    \node[inner sep=0pt, text width=.24\textwidth,execute at begin node=\tiny, below=0.5em of mnem1_1] (mnem1_2) {\color{orange}Aerial view of Bela Vista Park};
    \node[inner sep=0pt, text width=.24\textwidth,execute at begin node=\tiny, below=0.5em of mnem1_2] (mnem1_3) {\color{violet}Bela Vista Park};
    
    % \node[inner sep=0pt, text width=.24\textwidth,execute at begin node=\tiny, below=0.5em of img2] (tat2) {\color{blue}The park in 2007};
    \node[inner sep=0pt, text width=.24\textwidth,execute at begin node=\tiny, below=0.5em of img2] (mnem2_1) {\color{teal}Bela Vista Park};
    \node[inner sep=0pt, text width=.24\textwidth,execute at begin node=\tiny, below=0.5em of mnem2_1] (mnem2_2) {\color{orange}Aerial view of Bela Vista Park};
    \node[inner sep=0pt, text width=.24\textwidth,execute at begin node=\tiny, below=0.5em of mnem2_2] (mnem2_3) {\color{violet}Entrance to the park};
    
    % \node[inner sep=0pt, text width=.24\textwidth,execute at begin node=\tiny, below=0.5em of img3] (tat3) {\color{blue}The park in 2007};
    \node[inner sep=0pt, text width=.24\textwidth,execute at begin node=\tiny, below=0.5em of img3] (mnem3_1) {\color{teal}Bela Vista Park from space};
    \node[inner sep=0pt, text width=.24\textwidth,execute at begin node=\tiny, below=0.5em of mnem3_1] (mnem3_2) {\color{orange}Location of Bela Vista Park in Lisbon};
    \node[inner sep=0pt, text width=.24\textwidth,execute at begin node=\tiny, below=0.5em of mnem3_2] (mnem3_3) {\color{violet}Aerial view of Bela Vista Park};
    
    % \node[inner sep=0pt, text width=.24\textwidth,execute at begin node=\tiny, below=0.5em of img4] (tat4) {\color{blue}Aerial view of Bela Vista Park};
    \node[inner sep=0pt, text width=.24\textwidth,execute at begin node=\tiny, below=0.5em of img4] (mnem4_1) {\color{teal}Aerial view of the park};
    \node[inner sep=0pt, text width=.24\textwidth,execute at begin node=\tiny, below=0.5em of mnem4_1] (mnem4_2) {\color{orange}Aerial view of Bela Vista Park};
    \node[inner sep=0pt, text width=.24\textwidth,execute at begin node=\tiny, below=0.5em of mnem4_2] (mnem4_3) {\color{violet}Aerial view of Bela Vista Park};
    
  \end{tikzpicture}
  \\
    \bottomrule
 
    \end{tabular}
    \caption{\textbf{Qualitative samples of generated captions} with TAT-MNEM (top, {\color{teal}green}), GPT2-MNEM (middle, {\color{orange}orange}) and T5-MNEM (bottom, {\color{violet}purple}) for unpaired images given a similar Wikipedia article section. The models usually prefer to generate short, repetitive and template-based sentences. However, in some cases, they can integrate relevant visual details into the caption. 
    % The semantic relevance between image and section is given by a retrieval system~\cite{radford2021learning} where the semantic decreases from left to right.
    }
    \label{fig:story}
    \vspace{-0.5cm}
\end{figure*}

%% file: tables/nic_models.tex
\setlength{\tabcolsep}{4pt}
\begin{table*}
\tiny
\begin{center}
\caption{\textbf{Comparison between different transformer-based methods in News Image Captioning.} \checkmark means consisting of the component while \text{\sffamily X} means not. The components with (f) are frozen during training. *: excludes the data used to pre-train ResNet-152 and RoBERTa.}
\label{table:nic_models}
\begin{adjustbox}{width=1\textwidth}
\begin{tabular}{c|c|ccc|c|c|c}
\hline\noalign{\smallskip}
\multirow{2}{*}{Model} & \multirow{2}{*}{Method} & \multicolumn{3}{c|}{Model Architecture} & \multirow{2}{*}{Template} &\multirow{2}{*}{\#Trainable Params} & \multirow{2}{*}{Pre-training Data*}\\
\noalign{\smallskip}
\cline{3-5}
\noalign{\smallskip}
&  & Image Encoder & Text Encoder & Text Decoder &  &  & \\
\noalign{\smallskip}
\hline
\noalign{\smallskip}
TaT & End-to-end & ResNet-152(f) & RoBERTa(f) & Transformer & \text{\sffamily X} & 200M & \text{\sffamily X}\\
VisualNews & Two-stage & ResNet-152(f) & Transformer & Transformer & \checkmark & 93M & \text{\sffamily X}\\
JoGANIC & End-to-end & ResNet-152(f) & RoBERTa(f) & Transformer & \checkmark & 205M & Wikipedia KB\\
\noalign{\smallskip}
\hline
\noalign{\smallskip}
Ours & End-to-end & ResNet-152(f) & \text{\sffamily X} & Transformer & \text{\sffamily X} & 224M & WIT-2M\\
\noalign{\smallskip}
\hline
\end{tabular}
\end{adjustbox}
\end{center}
\vspace{-.4cm}
\end{table*}
\setlength{\tabcolsep}{1.4pt}

%% file: tables/wit_stats.tex
\setlength{\tabcolsep}{4pt}
\begin{table}
\tiny
\begin{center}
\caption{\textbf{Dataset statistics.} The split and statistics of the Wiki training data.}
\label{tab:wit_stats}
\begin{adjustbox}{width=1\columnwidth}
\begin{tabular}{lcccc}
\hline 
& Train & Val & Test & Total \\
\hline
Number of Samples & 2613732 & 8000 & 20000 & 2641732\\
Avg. Len. Caption (Words) & 8.87 & 8.71 & 8.91 & 8.88\\
Avg. Len. Description (Words) & 22.06 & 22.15 & 22.01 & 22.06 \\ 
Avg. Len. Section (Words) & 193.71 & 193.6 & 193.89 & 193.70\\ 
\#NEs in Captions (Word) & 34.36\% & 34.73\% & 34.94\% & 34.36\%\\
\#NEs in Descriptions (Word) & 32.14\% & 32.23\% & 32.57\% & 32.14\%\\
\#NEs in Sections (Word) & 22.06\% & 22.15\% & 22.12\% & 22.06\%\\
\#Captions with NEs & 84.99\% & 84.28\% & 84.1\% & 84.98\%\\
\#Descriptions with NEs & 92.1\% & 91.98\% & 91.6\% & 92.1\%\\
\#Sections with NEs & 98.58\% & 98.71\% & 98.56\% & 98.58\%\\
\hline
\end{tabular}
\end{adjustbox}
\end{center}
\vspace{-.4cm}
\end{table}
\setlength{\tabcolsep}{1.4pt}

%% file: figures/jaccard_dist.tex
\begin{figure*}[t!]
\centering
\begin{minipage}{.3\textwidth}
        \centering
        \includegraphics[width=.95\linewidth, height=0.2\textheight]{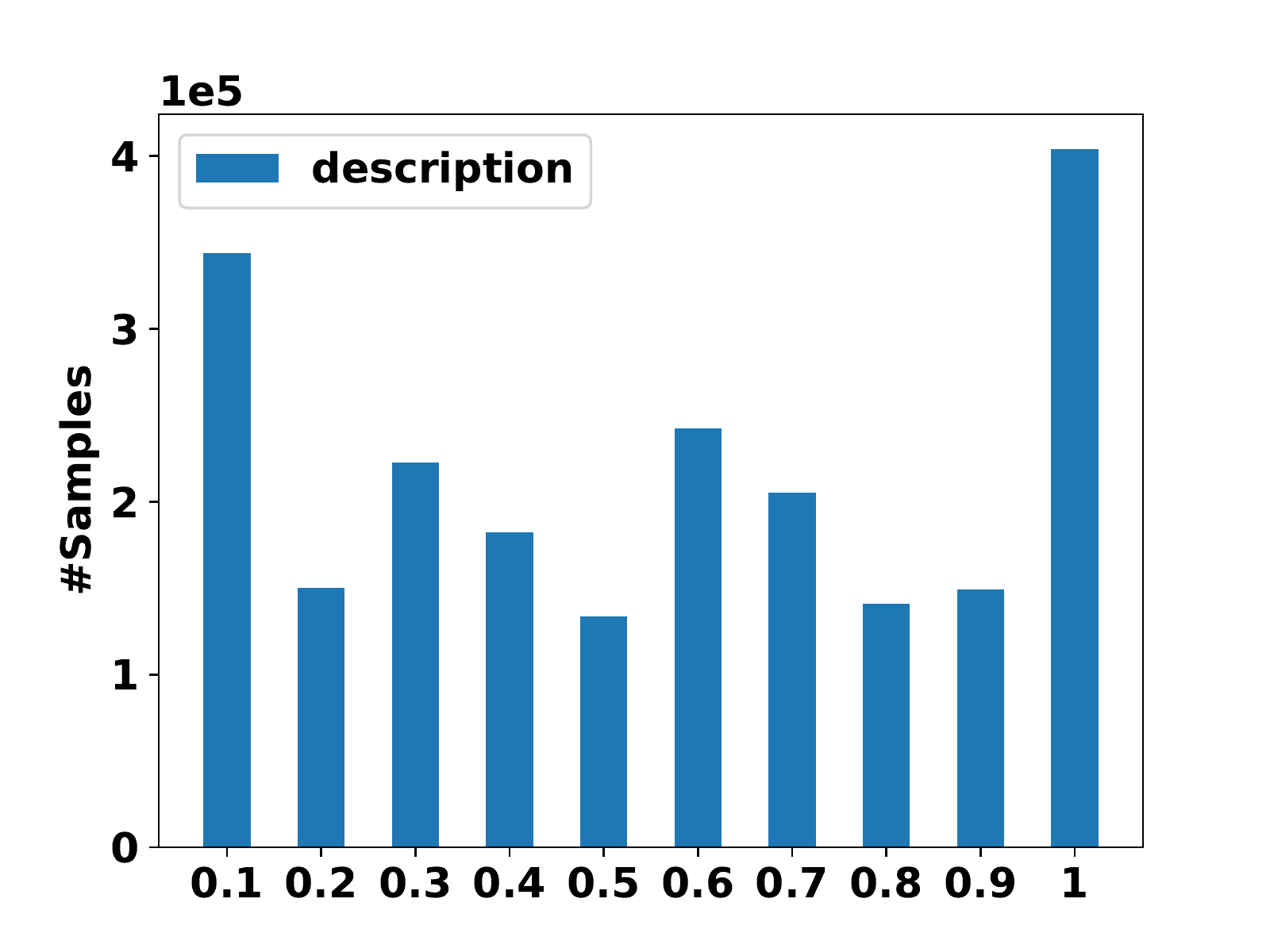}
        % \caption{Jaccard similarity of caption and description}
        % \label{fig:prob1_6_2}
    \end{minipage}%
    \begin{minipage}{0.3\textwidth}
        \centering
        \includegraphics[width=.95\linewidth, height=0.2\textheight]{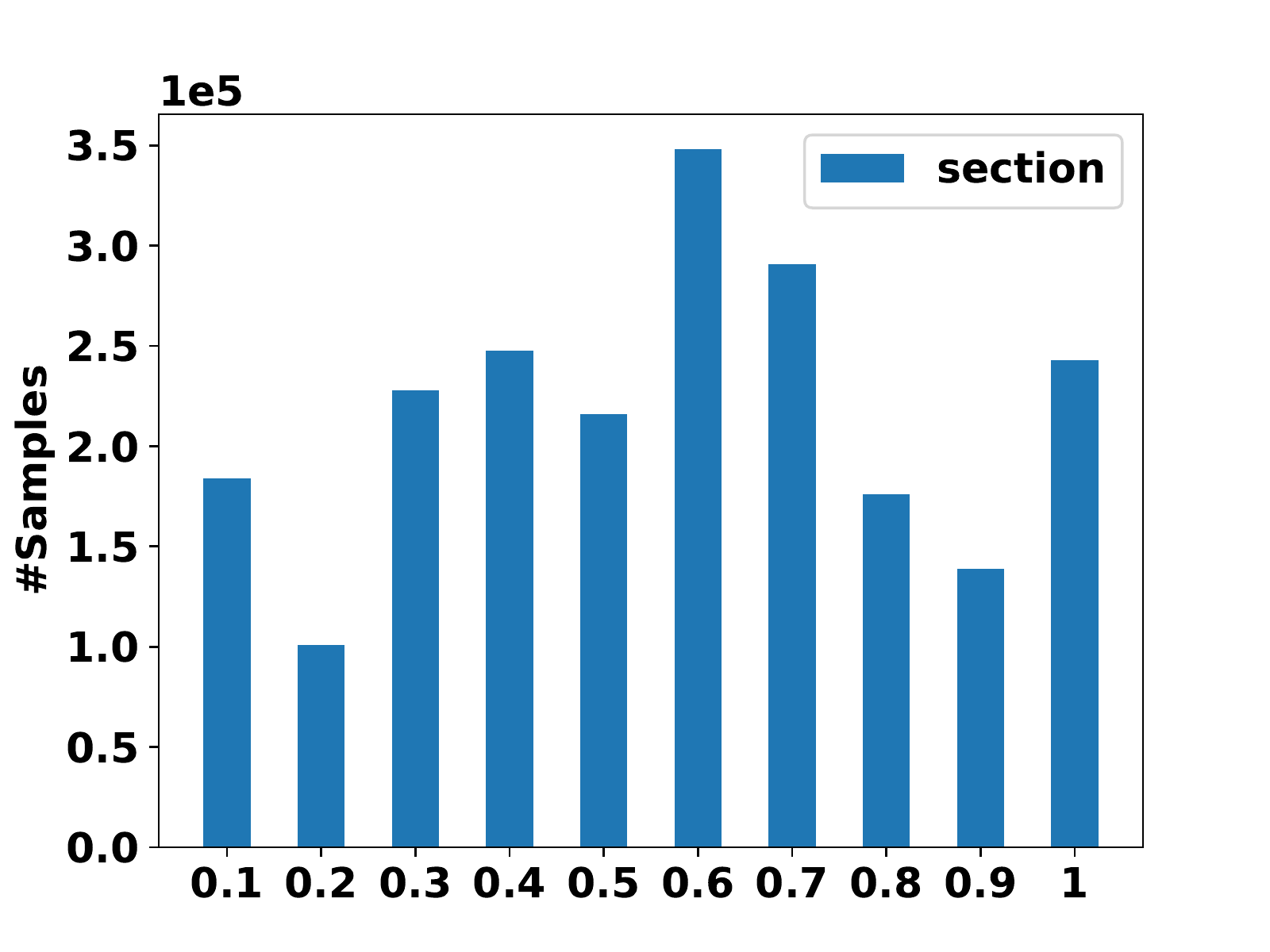}
        % \caption{Jaccard similarity of caption and section}
        % \label{fig:prob1_6_1}
    \end{minipage}
    \begin{minipage}{0.3\textwidth}
        \centering
        \includegraphics[width=.95\linewidth, height=0.2\textheight]{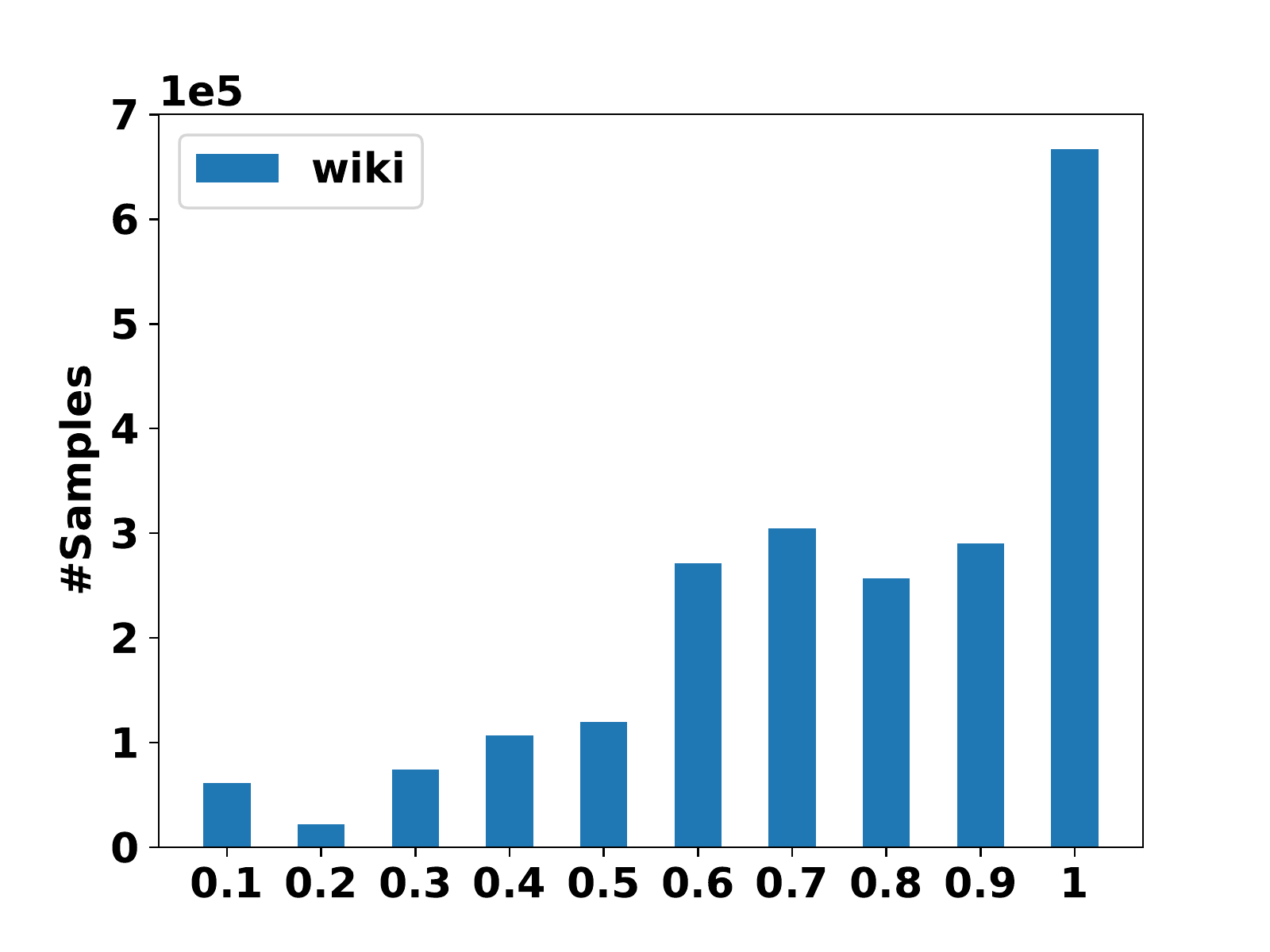}
        % \caption{Jaccard similarity of caption and description+section}
        % \label{fig:prob1_6_1}
    \end{minipage}
\caption{\textbf{Overlap between the caption and description (left); caption and section (right)}, measured by Jaccard similarity with a small modification. For each training example, we split each of these text into a sequence of words, then compute the intersection by adding up the length of all common words, then normalize it by the caption length. A score closer to 1 means a higher degree of intersection, i.e a larger chunk of the caption appears in the description. In this figure, each value $x$ in the horizontal axis indicates the interval $[x-1,x)$ except the last column, which is $[0.9,1]$}
\label{fig:jacc_dist}
\end{figure*}

%% file: figures/wit_v_gns.tex
\begin{figure}[t!]
\centering
    \includegraphics[width=\columnwidth, height=0.3\textheight]{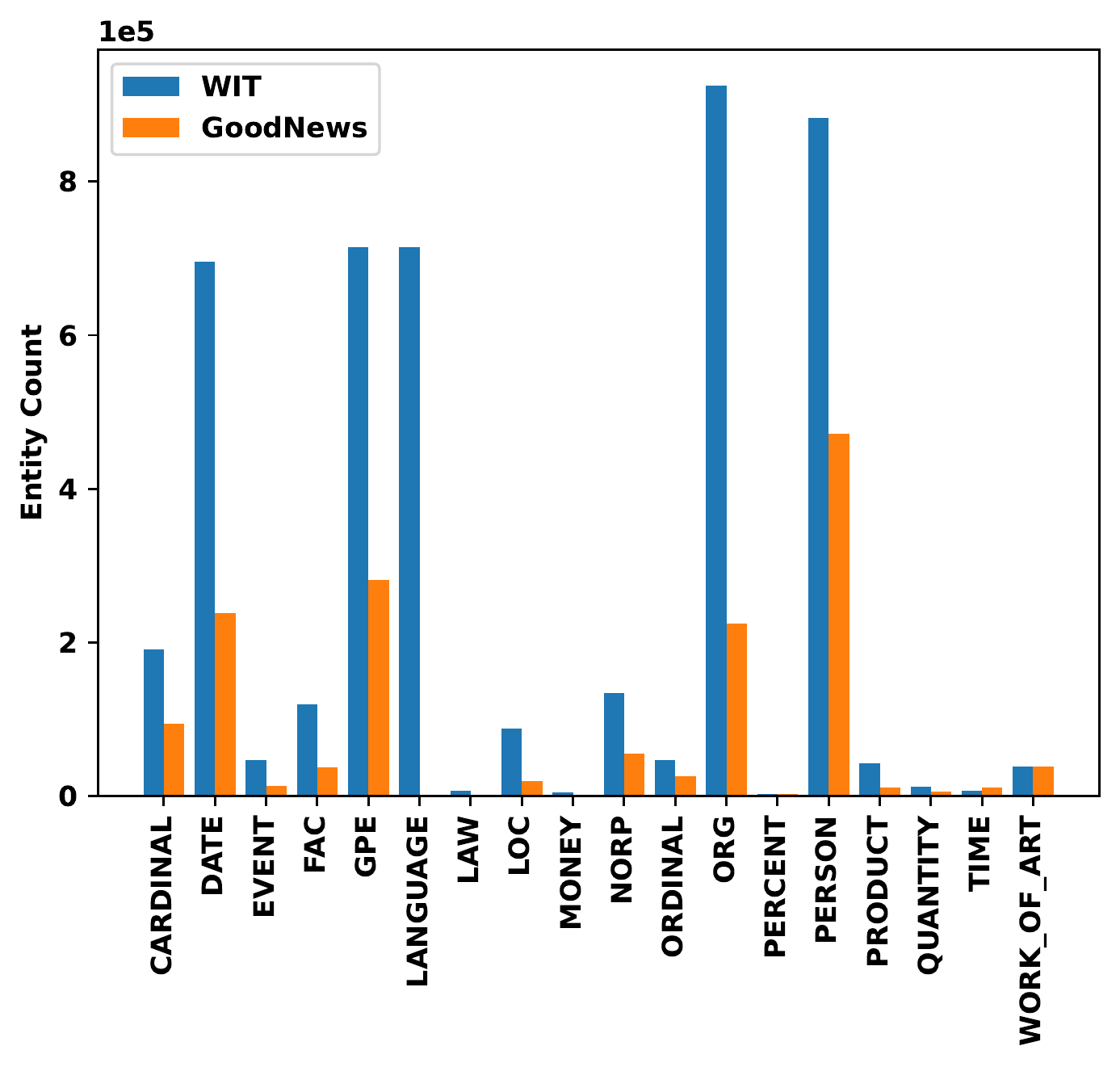}
\caption{\textbf{Entity categories and its cardinality in WIT and GoodNews}. Among most common types in both datasets are people’s names, organizations and geopolitical entities.}
\label{fig:wit_v_gns}
\end{figure}

%% file: tables/wit_vs_gns.tex
\setlength{\tabcolsep}{4pt}
\begin{table}
\tiny
\begin{center}
\caption{\textbf{Comparison} between WIT and GoodNews datasets.}
\label{tab:wit_vs_gns}
\begin{adjustbox}{width=1\columnwidth}
\begin{tabular}{lccc}
\hline 
& GoodNews & Wiki\\
\hline
Number of Samples & 466000 & 5411978\\
Unique Images & -- & 2082504\\ 
Unique (Image+Desc) & -- & 2550904\\
Unique (Image+Sec) & -- & 2640462\\
Average Caption Length (Words) & 18.21 & 8.88\\
Named Entities (Word) & 19.59\% & 35.31\%\\
Named Entities (Sentence) & 95.51\% & 96.29\%\\
Nouns & 46.70\% & 36.58\%\\
Verbs & 11.22\% & 8.03\%\\
Adjectives & 5\%  & 6.31\%\\
\hline
\end{tabular}
\end{adjustbox}
\end{center}
\vspace{-.4cm}
\end{table}
\setlength{\tabcolsep}{1.4pt}

%% file: tables/easy_vs_hard_context.tex
\setlength{\tabcolsep}{4pt}
\begin{table}[t!]
\tiny
\begin{center}
\caption{\textbf{Performance on Easy and Hard subsets.} Wiki. indicates the context combining both description and section. $\dagger$ indicates that MNEM pre-training is performed.}
\label{table:easy_v_hard_context}
\begin{adjustbox}{width=1\columnwidth}
\begin{tabular}{cccccccccc}
\hline\noalign{\smallskip}
Model & Context & Test Set & GT ol. & Gen. ol. & Gen. len. & B & C & Pr & Re\\
\hline
\parbox[t]{2mm}{\multirow{9}{*}{$\text{TaT}$}} & Desc. & Easy & 0.78 & 0.86 & 8.66 & 33.40 & 336.10 & 40.32 & 37.83\\
& Desc. & Hard & 0.19 & 0.64 & 6.65 & 7.06 & 95.27 & 16.55 & 13.57\\
& Desc. & Overall & 0.50 & 0.74 & 7.72 & 20.94 & 212.31 & 27.71 & 26.33\\
\cline{2-10}
& Sec. & Easy & 0.71 & 0.78 & 5.66 & 14.65 & 208.09 & 25.92 & 22.08\\
& Sec. & Hard & 0.25 & 0.61 & 5.49 & 4.40 & 76.23 & 12.50 & 8.50\\
& Sec. & Overall & 0.51 & 0.70 & 5.87 & 9.20 & 150.20 & 20.02 & 15.86\\
\cline{2-10}
& Wiki. & Easy & 0.81 & 0.92 & 7.65 & 29.93 & 313.99 & 39.49 & 37.2\\
& Wiki. & Hard & 0.35 & 0.75 & 6.03 & 8.48 & 133.68 & 19.06 & 16.06\\
& Wiki. & Full & 0.71 & 0.88 & 6.46 & \textbf{24.47} & \textbf{267.37} & \textbf{32.10} & \textbf{34.64}\\
\hline
\parbox[t]{2mm}{\multirow{9}{*}{$\text{TaT}^\dagger$}} & Desc. & Easy & 0.78 & 0.85 & 8.73 & 33.59 & 338.8 & 40.76 & 38.10\\
& Desc. & Hard & 0.19 & 0.62 & 6.72 & 7.28 & 98.33 & 16.88 & 13.84\\
& Desc. & Overall & 0.50 & 0.74 & 7.77 & 22.36 & 224.38 & 29.36 & 27.90\\
\cline{2-10}
& Sec. & Easy & 0.71 & 0.78 & 5.66 & 15.04 & 213.11 & 26.60 & 22.76\\
& Sec. & Hard & 0.25 & 0.60 & 6.22 & 4.79 & 78.18 & 13.13 & 8.94\\
& Sec. & Overall & 0.51 & 0.70 & 5.44 & 9.54 & 153.87 & 20.75 & 16.01\\
\cline{2-10}
& Wiki. & Easy & 0.81 & 0.92 & 7.89 & 30.76 & 319.01 & 39.71 & 38.20\\
& Wiki. & Hard & 0.35 & 0.73 & 6.16 & 8.91 & 137.81 & 19.39 & 16.53\\
& Wiki. & Full & 0.71 & 0.87 & 7.08 & \textbf{25.31} & \textbf{272.46} & \textbf{33.34} & \textbf{35.56}\\
\hline
\end{tabular}
\end{adjustbox}
\end{center}
\vspace{-0.5cm}
\end{table}
\setlength{\tabcolsep}{1.4pt}

%% file: tables/goodnews_ext.tex
\setlength{\tabcolsep}{4pt}
\begin{table}[t!]
\tiny
\begin{center}
\caption{\textbf{Performance of proposed models on GoodNews.} Ft means fine-tuning the models with the captioning objective in the corresponding dataset. Note that for captioning metrics, we directly use the results that are reported in the original papers.}
\label{table:goodnews_ext}
\begin{adjustbox}{width=1\columnwidth}
\begin{tabular}{c|cc|cccccc}
\hline\noalign{\smallskip}
\multirow{2}{*}{Model} & \multicolumn{2}{c|}{GoodNews} & \multirow{2}{*}{B} & \multirow{2}{*}{M} & \multirow{2}{*}{R} & \multirow{2}{*}{C} & \multirow{2}{*}{Pr} & \multirow{2}{*}{Re}\\
\noalign{\smallskip}
\cline{2-3}
\noalign{\smallskip}
& MNEM & Ft &  &  &  &  &  & \\
\noalign{\smallskip}
\hline
\noalign{\smallskip}	
\multirow{2}{*}{$\text{TaT}$} &  & $\checkmark$ & 6 & 10.1 & 21.2 & 53.1 & 21.3 & 19.7\\
& $\checkmark$ & $\checkmark$ & 6.23 & 10.08 & 22.47 & 55.14 & 19.8 & 18.72\\
\noalign{\smallskip}
\hline
\noalign{\smallskip}						
\multirow{2}{*}{$\text{T5}_{\textit{BASE}}$} &  & $\checkmark$ & 6.06 & 9.56 & 20 & 52.1 & 19.6 & 18.06\\
& $\checkmark$ & $\checkmark$ & 6.07 & 9.61 & 20.16 & 53.11 & 19.98 & 18.66\\ 
\noalign{\smallskip}
\hline
\end{tabular}
\end{adjustbox}
\end{center}
\vspace{-.7cm}
\end{table}
\setlength{\tabcolsep}{1.4pt}

%% file: figures/wiki_qual_results.tex
% 9613
\begin{table}[h!]
\tiny
\begin{tabular}{ c }
\toprule
\begin{tikzpicture}
  \tikzset{block/.append style={text width=0.25\textwidth, minimum height=0.2\textwidth, align=justify}}
  
    \node[inner sep=0pt] (img)     {\includegraphics[width=0.3\textwidth, height=0.2\textwidth]{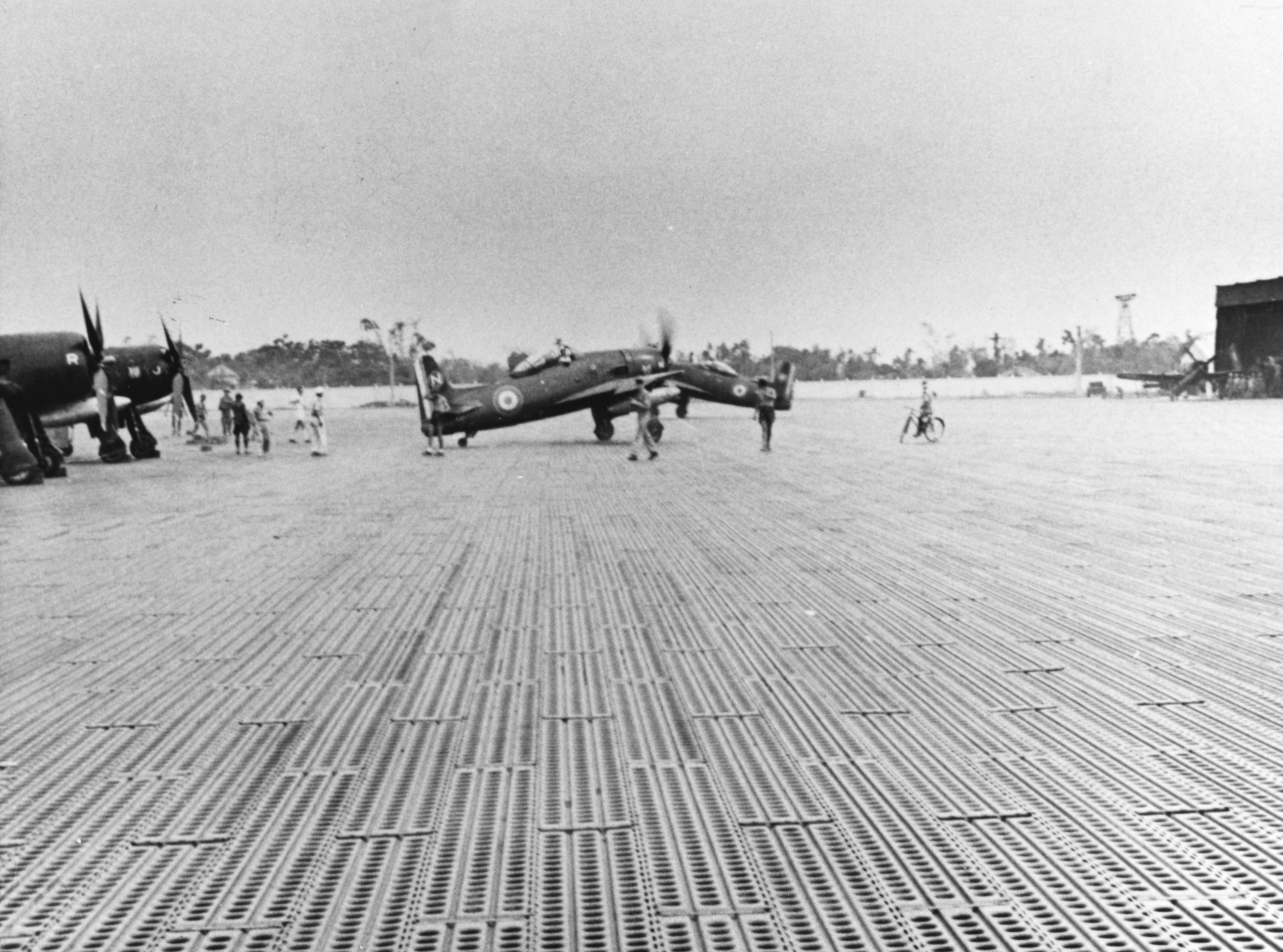}};
   
    \node[block, right=4.5em of img]                 (description) {\textbf{Description}\\ View of Tourane Airfield (Da Nang), Indochina, looking north, circa in 1954. Original caption: "New metal mat area on northeast side of runway. Two of three F8F aircraft observed departing on napalm mission. Aircraft returned in 35 minutes.".\vspace{8em}};
    \node[inner sep=0pt, above left=-0.90cm and 0.05cm of description] (logo) 
    {\includegraphics[width=.05\textwidth]{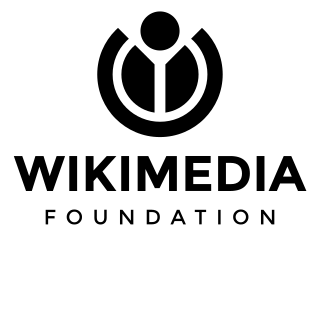}};
    
    \node[block, right=32em of img]                 (section) {\textbf{Section}\\ XF8F-1 Prototype aircraft, two built. F8F-1 Bearcat Single-seat fighter aircraft, equipped with folding wings, a retractable tailwheel, self-sealing fuel tanks, a very small dorsal fin, powered by a 2,100hp (1,566kW) Pratt and Whitney R-2800-34W Pratt \& Whitney R-2800-34W Double Wasp radial piston engine, armed with four 0.50 in (12.7 mm) machine guns, 658 built. ...\vspace{5em}};
    \node[inner sep=0pt, above left=-1.0cm and 0.1cm of section] (logo) 
    {\includegraphics[width=.05\textwidth]{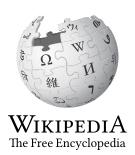}};
    
    \node[inner sep=0pt, execute at begin node=\tiny, align=left, below right=1em and -15.5em of img] (gt) {
    \textbf{Ground-Truth:} French Bearcats at Tourane Air Base, circa 1954.\\ \break
    \textbf{TaT:} F8F-1 in 1954 \\  \break
    \textbf{TaT+MNEM:} F8F-1 in 1954 \\ \break
    \textbf{GPT-2:} Tourane Airfield (Da Nang), Indochina, 1954\\  \break
    \textbf{GPT-2+MNEM:} Tourane Airfield, Indochina, 1954 \\  \break
    \textbf{T5:} Tourane Airfield (Da Nang), Indochina, looking north, circa 1954. \\ \break
    \textbf{T5+MNEM:} Two of three F8F aircraft observed departing on napalm mission at Tourane Airfield (Da Nang), Indochina, c. 1954};
    
\end{tikzpicture} \\
\bottomrule
\end{tabular}
\end{table}

% 3710
\begin{table}[h!]
\tiny
\begin{tabular}{ c }
\toprule
\begin{tikzpicture}
  \tikzset{block/.append style={text width=0.25\textwidth, minimum height=0.2\textwidth, execute at begin node=\tiny, align=justify}}
  
    \node[inner sep=0pt] (img)     {\includegraphics[width=0.3\textwidth, height=0.2\textwidth]{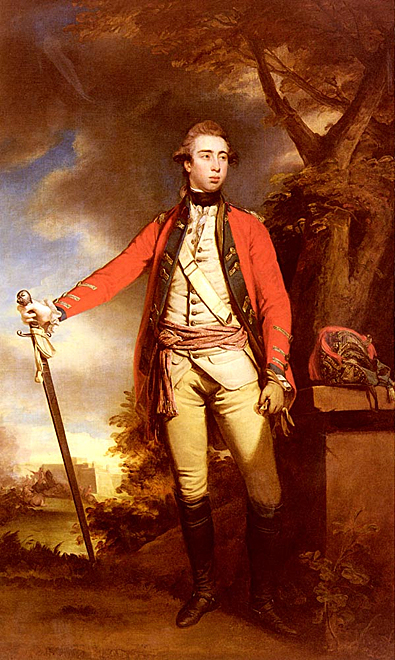}};
   
    \node[block, right=4.5em of img]                 (description) {\textbf{Description}\\ Portrait of George Townshend, Lord Ferrers, in uniform of the 15th Kings light Dragoons, standing full length.\vspace{10em}};
    \node[inner sep=0pt, above left=-0.90cm and 0.05cm of description] (logo) 
    {\includegraphics[width=.05\textwidth]{figures/wiki_qual_samples/Wikimedia_logo.png}};
    \node[block, right=32em of img]                 (section) {\textbf{Section}\\ George Townshend, 2nd Marquess Townshend, PC, FRS (18 April 1753 – 27 July 1811), known as The Lord Ferrers of Chartley from 1770 to 1784 and as The Earl of Leicester from 1784 to 1807, was a British peer and politician.\vspace{8em}};
    \node[inner sep=0pt, above left=-1.0cm and 0.1cm of section] (logo) 
    {\includegraphics[width=.05\textwidth]{figures/wiki_qual_samples/Wikipedia-logo.png}};
   
    \node[inner sep=0pt, execute at begin node=\tiny, align=left, below right=1em and -15.5em of img] (gt) {
    \textbf{Ground-Truth:} Lord Townshend by Sir Joshua Reynolds.\\ \break
    \textbf{TaT} George Townshend in uniform of the 15th Kings light Dragoons standing full length \\ \break
    \textbf{TaT+MNEM:} George Townshend, Lord Ferrers, in uniform of the 15th Kings light Dragoons standing full length \\ \break
    \textbf{GPT-2:} George Townshend, Lord Ferrers, in uniform of the 15th Kings light Dragoons, standing full length\\ \break
    \textbf{GPT-2+MNEM:} The Earl of Leicester \\ \break
    \textbf{T5:} George Townshend, 2nd Marquess Townshend \\ \break
    \textbf{T5+MNEM:} Lord Ferrers, in uniform of the 15th Kings Light Dragoons, standing full length};
    
  \end{tikzpicture} \\
\bottomrule
\end{tabular}
\end{table}
\clearpage
\newpage
% 6697
\begin{table}[h!]
\tiny
\begin{tabular}{ c }
\toprule
\begin{tikzpicture}
  \tikzset{block/.append style={text width=0.25\textwidth, minimum height=0.2\textwidth, execute at begin node=\tiny, align=justify}}
  
    \node[inner sep=0pt] (img)     {\includegraphics[width=0.3\textwidth, height=0.2\textwidth]{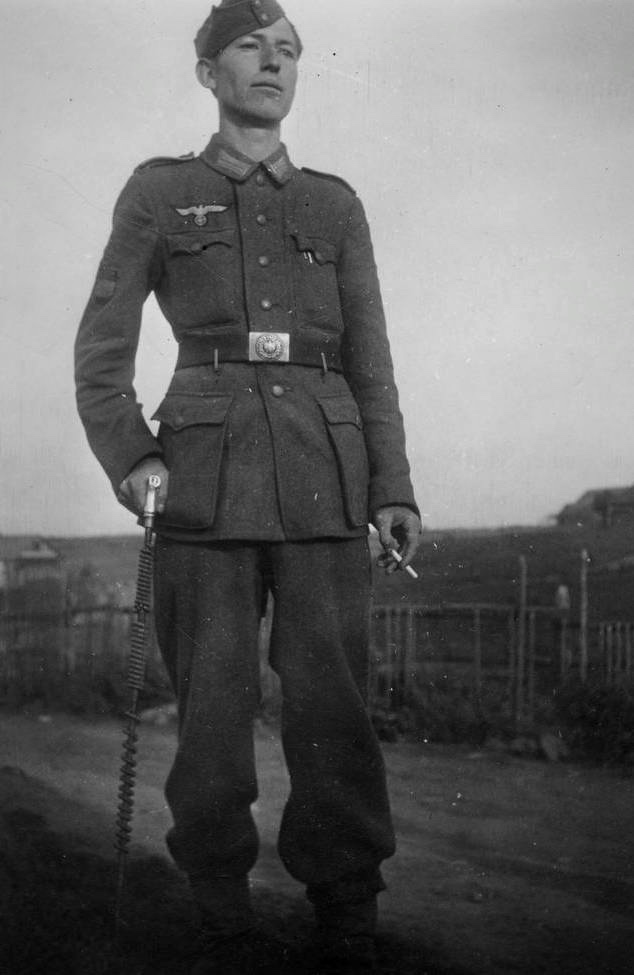}};
   
    \node[block, right=4.5em of img]                 (description) {\textbf{Description}\\ Español: Soldado no identificado de la División Azul española en el Frente Oriental.\vspace{11em}};
    \node[inner sep=0pt, above left=-0.90cm and 0.05cm of description] (logo) 
    {\includegraphics[width=.05\textwidth]{figures/wiki_qual_samples/Wikimedia_logo.png}};
    
    \node[block, right=32em of img]                 (section) {\textbf{Section}\\ Devised in June 1943, Nurnberg was purely a defensive operation in the Pyrenees along both sides of the Spanish-French border in the event of Allied landings in the Iberian peninsula, which were to repel an Allied advance from Spain into France.\vspace{8em}};
    \node[inner sep=0pt, above left=-1.0cm and 0.1cm of section] (logo) 
    {\includegraphics[width=.05\textwidth]{figures/wiki_qual_samples/Wikipedia-logo.png}};
   
    \node[inner sep=0pt, execute at begin node=\tiny, align=left, below right=1em and -15.5em of img] (gt) {
    \textbf{Ground-Truth:} A Spanish volunteer of the Blue Division\\ \break
    \textbf{TaT} A Spanish soldier in the Pyrenees \\ \break
    \textbf{TaT+MNEM:} Spanish Soldier of the Division Azul in the Pyrenees \\ \break
    \textbf{GPT-2:} Nurnberg in 1943 \\ \break
    \textbf{GPT-2+MNEM:} Luis Nurnberg \\ \break
    \textbf{T5:} Azul division in the Frente Oriental. \\ \break
    \textbf{T5+MNEM:} Nurnberg in the Spanish-French border.};
  \end{tikzpicture} \\
\bottomrule
\end{tabular}
\end{table}